\documentclass[11pt]{interact}
\usepackage{hyperref}

\usepackage[utf8]{inputenc}
\usepackage{enumitem}
\usepackage[table]{xcolor}
\usepackage{hyperref}
\hypersetup{
    colorlinks=true,
    linkcolor=purple,
    citecolor=blue,
    filecolor=magenta,      
    urlcolor=cyan,
}
\usepackage{multirow}
\usepackage{float}
\usepackage[titletoc]{appendix}
\usepackage[font=small]{caption}
\usepackage{subcaption}
\usepackage{graphicx}
\usepackage{geometry}
\usepackage{booktabs}   
\usepackage{pifont}     
\usepackage{adjustbox}  
\usepackage{tabularx}
\usepackage{array}
\usepackage[sort,authoryear]{natbib}
\bibpunct[, ]{(}{)}{;}{a}{}{,}

\theoremstyle{plain}

\usepackage[most]{tcolorbox}

\usepackage{courier}

\newcommand{\rev}[1]{{\color{black}{#1}}} 

\usepackage{csquotes}
\MakeOuterQuote{"}

\usepackage{caption}

\articletype{Research Article}

\begin{document}

\title{Automating Supply Chain Disruption Monitoring via an Agentic AI Approach}
\author{
\name{Sara AlMahri\textsuperscript{a}\thanks{CONTACT Sara AlMahri. Email: sa2104@cam.ac.uk}, Liming Xu\textsuperscript{a},  and Alexandra Brintrup\textsuperscript{a,}\textsuperscript{b}}
\affil{\textsuperscript{a}Institute for Manufacturing, Department of Engineering, University of Cambridge, Cambridge, UK}
\affil{\textsuperscript{b}The Alan Turing Institute, London, UK}}
\maketitle 

\maketitle 
\begin{abstract}
Modern supply chains are increasingly exposed to disruptions from geopolitical events, demand shocks, trade restrictions, to natural disasters. 
While many of these disruptions originate deep in the supply network, most companies still lack visibility beyond Tier-1 suppliers, leaving upstream vulnerabilities undetected until the impact cascades downstream. 
To overcome this blind-spot and move from reactive recovery to proactive resilience, we introduce a minimally supervised agentic AI framework that autonomously monitors, analyses, and responds to disruptions across extended supply networks. The architecture comprises seven specialised agents powered by large language models and deterministic tools that jointly detect disruption signals from unstructured news, map them to multi-tier supplier networks, evaluate exposure based on network structure, and recommend mitigations such as alternative sourcing options.
\rev{We evaluate the framework across 30 synthesised scenarios covering three automotive manufacturers and five disruption classes. The system achieves high accuracy across core tasks, with F1 scores between 0.962 and 0.991, and performs full end-to-end analyses in a mean of 3.83 minutes at a cost of \$0.0836 per disruption. Relative to industry benchmarks of multi-day, analyst-driven assessments, this represents a reduction of more than three orders of magnitude in response time. A real-world case study of the 2022 Russia--Ukraine conflict further demonstrates operational applicability. This work establishes a foundational step toward building resilient, proactive, and autonomous supply chains capable of managing disruptions across deep-tier networks.}
\end{abstract}


\section{Introduction}\label{sec:introduction}

Global supply chains have never been more vulnerable. In recent years, a series of major disruptions, from geopolitical tensions and demand shocks to trade restrictions and natural disasters, have exposed both the fragility and the deeply interconnected nature of modern supply networks. The US--China trade tensions led to tariffs on approximately \$360 billion of Chinese imports \citep{fajgelbaum2022economic}, forcing companies to restructure supply chains, diversify suppliers, and absorb increased operational costs. China's export controls on gallium and germanium, materials critical to semiconductor and defense industries, further strained global supply chains. Estimates suggest that companies may lose up to 40\% of one year's profits over a decade due to such disruptions \citep{mckinsey2020resilience}. These are not isolated incidents. They represent a new reality in which supply chain disruption management has become a strategic imperative.

\rev{Yet, a critical pattern emerges when examining these disruptions: many originate not from direct suppliers, but from deep within extended supply networks.} Recent studies have shown that over one-third, and in some cases up to half, of supply chain disruptions emerge beyond Tier-1 suppliers, within the deeper layers of the network \citep{berger2023risk}. A factory fire at a sub-tier semiconductor supplier, a regulatory change affecting a raw material producer, a labour strike at a logistics hub serving multiple upstream manufacturers: these events often go undetected until they cascade downstream and reach immediate suppliers. By then, the damage is already propagating through the network. \rev{This raises a fundamental question: if disruptions originate deep in extended supply networks, why do most monitoring solutions focus only on direct suppliers?}

\rev{The answer lies in the challenge of extended supply chain visibility. Mapping supplier relationships beyond Tier-1 is inherently difficult: networks span thousands of entities across multiple continents, relationships shift as suppliers merge, fail, or restructure, and data remains fragmented across disparate systems. In response to growing disruption risks, companies have adopted digital solutions to improve visibility and resilience. Yet these solutions remain constrained to direct supplier relationships precisely because multi-tier mapping at scale has proven technically and operationally prohibitive. \textit{Enterprise visibility platforms} integrate real-time data on inventory, shipments, and supplier activity, providing dashboards that track operational status \citep{ivanov2021digital}. However, these platforms monitor only direct supplier relationships, leaving upstream tiers invisible. \textit{Supply chain control towers} offer centralized oversight and predictive analytics across global operations, triggering alerts when predefined thresholds are breached \citep{bartlett2007improving}. Yet, control towers operate on structured, internal data and predefined rules for Tier-1 suppliers only. They cannot process emerging signals from external sources or detect anomalies in upstream tiers. \textit{Digital supply chain twins} simulate end-to-end operations using telemetry and scenario analysis, supporting what-if planning and policy evaluation \citep{longo2023digital, roman2025digitaltwins}. While powerful for experimentation, digital twins are typically instantiated around Tier-1 and internal operations, leaving blind spots in deep tiers where early disruption cues often surface.

\rev{Researchers have proposed complementary approaches, but each carries fundamental limitations.} For instance, \textit{Graph-based risk propagation models} quantify how shocks cascade over inter-firm networks, estimating amplification pathways and systemic exposure \citep{tabachova2024estimating, xie2023assessment, sun2025risk}. These models provide formal propagation metrics and counterfactual stress tests. However, they require granular, structured network data as input. They cannot process emerging, unstructured signals such as breaking news or regulatory notices. \rev{Yet this is precisely where disruption intelligence should begin: detecting events as they unfold and mapping them directly onto extended supplier networks to identify which specific chains are exposed.} \textit{Traditional multi-agent systems (MAS)} distribute functionality across specialised agents executing predefined tasks, enabling decentralized detection and mitigation of localized disruptions \citep{wooldridge2009introduction, ferber1999multi, GiannakisLouis2011, bi2022model}. However, MAS implementations rely on hard-coded decision logic and static ontologies. They require manual encoding of each scenario, leading to brittle performance when facing novel disruptions that fall outside predefined rules \citep{bi2022model}.}

\rev{Despite the diversity of these technologies, they share a common limitation: none provides extended supply chain visibility beyond Tier-1, yet over one-third of disruptions originate beyond Tier-1.} This fundamental mismatch between where disruptions occur and where monitoring systems focus represents a critical gap in supply chain risk management.

\rev{This brings us to the core operational problem. Supply chain resilience, measured by time-to-survive (TTS) and time-to-recover (TTR) \citep{simchi2014superstorms, ivanov2017literature}, requires early detection and timely action. But while companies can only control their direct suppliers, disruption signals are scattered across unstructured sources (news articles, regulatory filings, supplier advisories) that are siloed, disconnected from network data, and provide no actionable insight until mapped to operations. \rev{The complete capability required is thus: (1) capturing disruption signals from scattered unstructured sources, (2) mapping them onto the company's extended supplier network, (3) tracing how disruptions propagate toward Tier-1 suppliers, (4) quantifying risk for the suppliers companies can control, and (5) deriving data-driven decisions to act before cascading effects materialize. No existing system performs this end-to-end process autonomously.}}

\rev{Today, supply chain managers attempt this process manually.} They read news feeds, scan supplier emails, and maintain ad-hoc spreadsheets to track potential risks. According to a recent industry survey \citep{deloitte2020outsourcing}, only 40\% of companies use dedicated tools to systematically log and report disruptions. But manual monitoring is infeasible at scale. A typical automotive manufacturer has hundreds of Tier-1 suppliers, thousands of Tier-2 suppliers, and tens of thousands of entities across deeper tiers. Continuously monitoring unstructured sources, mapping disruptions to this network, tracing propagation paths, quantifying risk, and deriving decisions requires an autonomous system.

\rev{Agentic AI provides a theoretical foundation for building such an autonomous system.} We define agentic AI as agents driven by large language models (LLMs) that plan, use tools, and coordinate via language. This is distinct from traditional multi-agent systems composed of rule-based or policy-based software agents \citep{wooldridge2009introduction, ferber1999multi}, which rely on hard-coded decision logic and static ontologies that limit adaptability to novel disruptions. \rev{In contrast, agentic AI leverages LLMs that can autonomously decompose high-level objectives into reasoning chains, interpret unstructured inputs, and adaptively generate sub-plans in real time \citep{brown2020language, bommasani2021opportunities, yao2023tree}.} Whereas MAS agents require manual specification of interaction schemas and scenario libraries, agentic architectures embed prompting strategies and internal planning loops. This enables dynamic goal formulation and flexible workflow reconfiguration without bespoke coding.

\rev{For supply chain disruption monitoring, agentic AI is theoretically well-suited for three reasons.} First, disruptions often first appear in unstructured sources (news articles, regulatory filings, supplier advisories) that require natural language understanding, a core capability of LLMs. Second, the multi-tier nature of supply networks requires dynamic reasoning about complex relationships that cannot be fully pre-encoded in static rules. Third, the need to bridge unstructured information processing with structured network analysis requires a system that can orchestrate multiple capabilities, from text extraction to graph traversal to risk calculation, autonomously.

This paper introduces the first minimally supervised agentic framework for end-to-end supply chain disruption monitoring, assessment, and mitigation. This framework is the first to perform the complete transformation: detect disruptions from scattered unstructured sources, map them to extended supplier networks across multiple tiers, trace propagation paths toward Tier-1, quantify risk based on network structure, and derive actionable decisions. By doing so, the framework enables a critical shift in supply chain strategy: from reactive recovery to proactive detection and reconfiguration.

\rev{However, deploying LLM-powered agents in critical supply chain decision-making contexts requires addressing well-documented risks. LLMs are known to hallucinate, generating factually incorrect information with apparent confidence. They may over-rely on training data patterns that do not generalise to novel disruptions. In high-stakes domains like supply chain management, such failures could lead to missed disruptions or false alarms.}

\rev{We address these risks through three complementary mechanisms grounded in established AI safety paradigms. First, \textit{retrieval-augmented grounding}: all factual claims, including company names, supplier relationships, and disruption details, are verified against the knowledge graph rather than generated from LLM training data alone. Second, \textit{deterministic tool orchestration}: critical computations, including risk score calculations, graph traversals, and supplier matching, are delegated to deterministic functions. LLMs are used primarily for reasoning, interpretation, and tool selection rather than direct computation. Third, \textit{human-in-the-loop oversight}: final review checkpoints allow supply chain managers to approve, revise, or override agent recommendations before execution. \rev{This follows established paradigms for AI-assisted decision support in high-stakes domains, ensuring that the system operates as a minimally supervised framework that enhances rather than replaces human expertise.}}

Below, we summarize main contributions of this work are as follows:

\begin{itemize}
    \item \rev{We present the first minimally-supervised agentic, LLM-enabled framework for supply-chain disruption monitoring and mitigation across multi-tier networks. The framework autonomously transforms unstructured disruption signals into mapped network knowledge and actionable intelligence, addressing the fundamental gap between where disruptions are reported (unstructured text) and where impacts must be assessed (extended supplier networks).}
    
    \item We present a set of modular, task-specific prompts, designed using chain-of-thought reasoning to guide LLM agents in performing key supply chain tasks, including disruption classification, multi-tier risk assessment, and supplier reconfiguration analysis.
    
    \item \rev{We introduce the first comprehensive synthesized evaluation dataset for supply chain disruption monitoring, comprising 30 manually synthesized scenarios across three automotive manufacturers, providing a benchmark for future research in this domain.}
\end{itemize}

\rev{The remainder of the paper is structured as follows.} \autoref{sec:related_work} reviews existing literature on supply chain disruption management.  \autoref{sec:framework} describes the framework architecture and agent design. \rev{\autoref{sec:evaluation} presents experimental evaluation across 30 scenarios, including methodology, performance metrics, and results discussion.} \rev{\autoref{section:case-study} demonstrates the framework's application using a real-world disruption scenario, illustrating operational value for supply chain managers.} \rev{\autoref{sec:discussion} addresses industry prerequisites and deployment considerations.} \rev{Finally, \autoref{sec:conclusions} summarizes contributions and outlines future research directions.}

\section{Related Work}
\label{sec:related_work}

\rev{This section reviews prior work across three research streams relevant to autonomous supply chain disruption monitoring: (1) network-based risk propagation models, (2) multi-agent systems for supply chain coordination, and (3) large language model-powered systems for supply chain disruption. For each stream, we examine technical capabilities, architectural constraints, and implications for autonomous disruption monitoring.}

\subsection{\rev{Network-Based Risk Propagation Models}}

\rev{Network-based approaches model supply chains as graphs where nodes represent firms and edges represent supplier-customer relationships. \citet{craighead2007severity} established that network topology, specifically density, node criticality, and structural complexity, determines how severely disruptions impact system performance. \citet{kim2015supply} demonstrated that supply networks exhibit scale-free properties: highly connected hub firms create systemic vulnerabilities where single-node failures cascade disproportionately. \citet{brintrup2018predicting} showed that centrality metrics (degree, betweenness, PageRank) can identify which nodes' failure would maximally impact the network.

Building on these structural insights, researchers developed formal propagation models. \citet{tabachova2024estimating} created graph-based cascade models that estimate disruption amplification across supplier networks using epidemic-style propagation dynamics (analogous to SIS/SIR models). \citet{xie2023assessment} proposed centrality-weighted risk scores that aggregate exposure across network paths. \citet{sun2025risk} extended these models to multi-tier networks, quantifying how disruptions at Tier-3 or Tier-4 amplify as they propagate toward the focal firm.

A critical finding motivates our focus on extended networks: \citet{berger2023risk} showed that over one-third of disruptions emerge beyond Tier-1 suppliers. This has driven research on multi-tier supply chain mapping \citep{almahri2024enhancing, musa2014supply, busse2017extending}, providing visibility infrastructure that enables deep-tier risk analysis.

These network models provide rigorous quantitative frameworks for risk assessment. However, they share a fundamental architectural constraint: they require structured inputs. Specifically, they take as input: (1) a pre-defined network adjacency matrix, (2) specification of which node(s) are disrupted, and (3) disruption magnitude or probability parameters. They then compute propagation effects across the network. They cannot process unstructured text such as news articles or regulatory filings. They propagate \textit{known} disruptions but cannot detect \textit{new} ones. The detection stage, identifying from external sources that a disruption has occurred and which entities are affected, must be performed manually before these models can be applied.

\subsection{\rev{Multi-Agent Systems for Supply Chain Coordination}}

Multi-Agent Systems (MAS) distribute decision-making across autonomous software agents that communicate via explicit protocols \citep{wooldridge2009introduction}. In supply chain applications, agents typically represent supply chain entities (suppliers, manufacturers, distributors) and coordinate through message-passing using protocols such as FIPA-ACL or KQML \citep{ferber1999multi}. Each agent operates according to predefined behavioural rules encoded in its decision logic.

Early MAS research focused on coordination and simulation. \citet{Swaminathan1998Modeling} represented supply chain entities as agents to simulate material and information flows. \citet{Caridi2005Improving} applied MAS to Collaborative Planning, Forecasting, and Replenishment (CPFR), enabling partner collaboration through predefined interaction protocols. \citet{LeeKim2008} reviewed MAS potential for real-time supply chain simulation, noting scalability challenges as agent numbers increase.

Subsequent research incorporated learning within MAS architectures. \citet{Brintrup2010Behaviour} demonstrated role-specific reinforcement learning enabling agents to reallocate tasks dynamically. \citet{Zhang2020Multi} applied multi-agent reinforcement learning to vehicle routing. Applications expanded to supplier selection \citep{Ameri2013Multi, Ghadimi2018}, procurement automation \citep{Xu2024Implementing}, and contract negotiation \citep{Jiao2006}. For disruption management specifically, \citet{GiannakisLouis2011} developed a multi-agent decision support system for manufacturing risk visibility, while \citet{BiEtAl2024} proposed heterogeneous risk management incorporating agent risk attitudes. For comprehensive surveys, see \citet{Xu2021Will}.

Despite these advances, traditional MAS exhibit architectural constraints that limit their applicability to disruption monitoring. First, agents operate on static ontologies: concepts, relationships, and valid actions must be explicitly defined at design time \citep{Holland:1986b}. Agents cannot interpret situations outside their encoded knowledge structures. Second, decision logic is rule-based: agents select actions based on if-then conditions that must anticipate every relevant scenario \citep{Lee1998}. Novel disruption types not encoded in rules cannot be handled. Third, agents cannot process unstructured natural language: news articles, regulatory filings, and supplier advisories must be manually parsed and converted to structured formats before agents can act \citep{busoniu2008comprehensive}. These constraints mean that while MAS can coordinate responses once disruptions are detected and structured, they cannot perform the initial detection from unstructured external sources.

\subsection{\rev{Large Language Model-Powered Supply Chain Systems}}

Large language models (LLMs) enable a fundamentally different approach: agents powered by neural language models that can process unstructured text, reason about novel situations, and generate contextual responses without explicit rule encoding \citep{achiam2023gpt, kojima2022large}. This capability addresses the core limitation of traditional MAS, the inability to interpret unstructured information where disruption signals first appear.

Recent work has applied LLM-powered agents to supply chain problems. SHIELD \citep{cheng2024shield} used LLMs to infer disruption-relevant schema from text for early warning, demonstrating that language models can extract supply chain concepts from unstructured sources. However, SHIELD operates as a standalone inference system without multi-agent coordination or network mapping capabilities. \citet{quan2024invagent} developed InvAgent, deploying dialogue-driven LLM agents for inventory management tasks including demand forecasting, safety-stock calculation, and replenishment ordering. \citet{li2024optimizing} combined LLM planning and negotiation agents to optimise order quantities and delivery schedules. \citet{jannelli2024agentic} created an agentic procurement framework where LLM agents orchestrate supplier evaluation and selection.

These systems demonstrate that LLM-powered agents can reason about supply chain problems and coordinate complex decisions. However, they share a common architectural pattern: they address \textit{downstream} supply chain tasks that assume disruption context is already known. InvAgent manages inventory given demand signals; it does not detect disruptions from news. The procurement framework selects suppliers; it does not monitor for supplier-affecting events. The planning agents optimise schedules; they do not identify schedule-disrupting incidents. None of these systems perform autonomous monitoring of external sources, extraction of disruption entities from unstructured text, or mapping of detected events to extended supplier networks.

Additionally, deploying LLMs in operational contexts requires addressing well-documented technical risks. Hallucinations occur when models generate factually incorrect statements with apparent confidence \citep{ji2023hallucination}. Prompt sensitivity causes outputs to vary significantly with small input changes \citep{liu2023lostmiddle, liang2022helm}. Limited grounding means base models do not access current information or cite authoritative sources \citep{lewis2020rag}. These risks necessitate architectural safeguards including retrieval-augmented generation, deterministic tool execution for critical computations, and human oversight for high-stakes decisions.

\subsection{\rev{Research Gap and Positioning}}

Table~\ref{tab:literature_synthesis} summarizes how each research stream addresses capabilities required for autonomous disruption monitoring.

\begin{table}[h]
\centering
\caption{\rev{Existing Approaches vs. Requirements for Autonomous Disruption Monitoring}}
\label{tab:literature_synthesis}
\resizebox{\textwidth}{!}{%
\begin{tabular}{lcccc}
\toprule
\textbf{Requirement} & \textbf{Network Models} & \textbf{Traditional MAS} & \textbf{LLM Systems} & \textbf{Our Framework} \\
\midrule
Process unstructured text (news, filings) & \ding{55} & \ding{55} & \ding{51} & \ding{51} \\
Detect emerging disruptions autonomously & \ding{55} & \ding{55} & \ding{55} & \ding{51} \\
Map to extended multi-tier networks & \ding{51} & \ding{55} & \ding{55} & \ding{51} \\
Quantify risk based on network structure & \ding{51} & \ding{55} & \ding{55} & \ding{51} \\
Adapt to novel disruption types & \ding{55} & \ding{55} & \ding{51} & \ding{51} \\
Derive actionable decisions & \ding{55} & \ding{51} & \ding{51} & \ding{51} \\
End-to-end autonomous pipeline & \ding{55} & \ding{55} & \ding{55} & \ding{51} \\
\bottomrule
\end{tabular}%
}
\end{table}

Network-based models provide rigorous propagation analysis and multi-tier visibility but require pre-structured inputs specifying which entities are disrupted. Traditional MAS provide coordination frameworks but rely on static ontologies and rule-based logic that cannot process unstructured text or adapt to novel situations. Existing LLM-powered systems demonstrate flexible reasoning over text but focus on downstream optimisation tasks (inventory, procurement, scheduling) rather than disruption detection and network mapping.

No existing system performs the complete transformation required for proactive disruption management: (1) detect disruptions from unstructured external sources, (2) map detected events to extended multi-tier supplier networks, (3) quantify risk propagation based on network structure, and (4) derive actionable decisions for supply chain managers. This requires integrating capabilities across all three streams: LLM reasoning for unstructured text processing, graph-based analysis for network propagation, and multi-agent coordination for decision-making.

Our framework addresses this gap by combining LLM-powered agents with deterministic graph tools and structured communication protocols. Unlike traditional MAS, agents interpret unstructured text through language model reasoning. Unlike standalone LLM systems, agents orchestrate graph database queries to map detected entities to extended supplier networks. Unlike network models, the framework autonomously detects disruptions from external sources rather than requiring pre-structured inputs. The following section describes the architecture that realises this integration.}

\section{Agentic Supply Chain Disruption Monitoring Framework}
\label{sec:framework}

This section introduces the architecture of our agentic framework for monitoring supply chain disruptions using LLM-powered agents. {We outline the overall structure and sequential stages of the disruption monitoring system.} {We introduce the concept of an agent and its fundamental components.} {We then describe the system architecture, emphasising how agents are organised and orchestrated to form an integrated disruption monitoring pipeline.} {We detail the roles of each agent, their reasoning tasks, tool integrations, and outputs.} {We discuss prompt engineering strategies that shape agent behaviour.} {We describe standardised communication protocols that ensure reliable agent interaction.} {Finally, we address system configuration and modularity considerations.} This section provides the technical foundation for understanding the framework's capabilities and design principles.

\subsection{Framework Overview}
\label{subsec:framework-overview}
\begin{figure}[H]
\centering
    \includegraphics[width=1\linewidth]{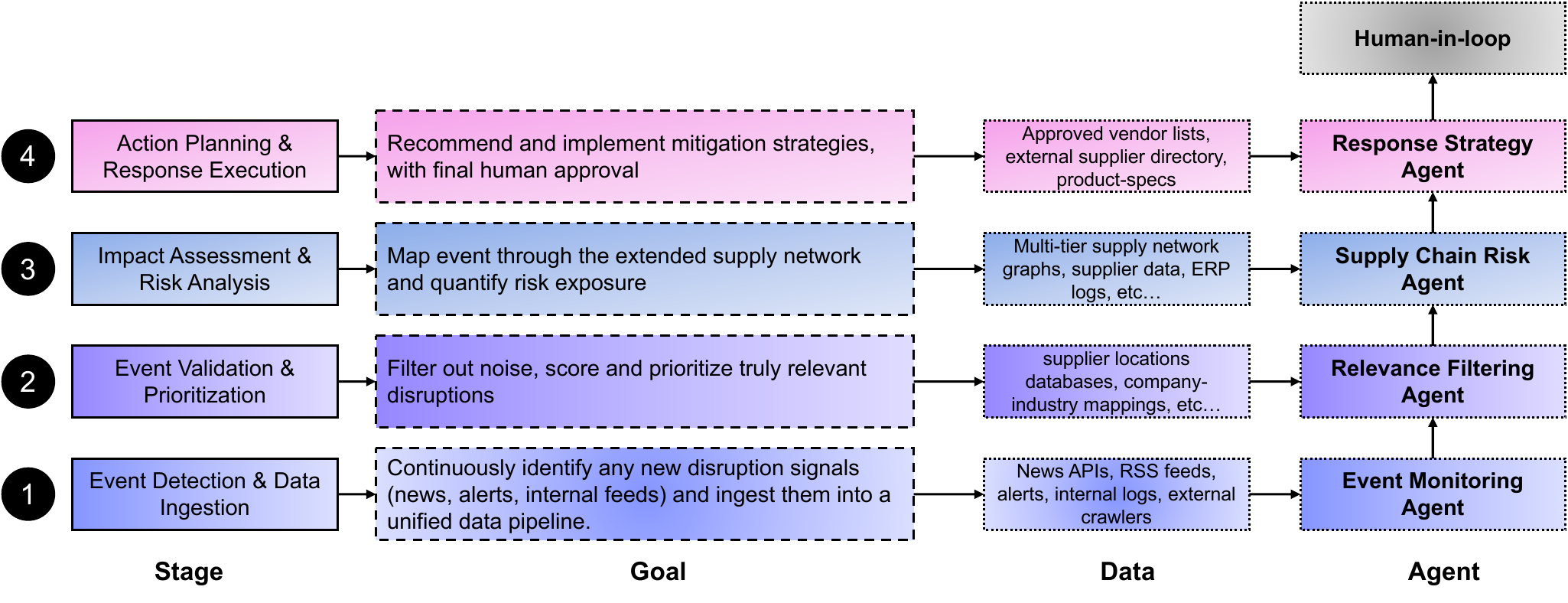}
    \caption{Overall stages, goals and data needed for autonomous disruption monitoring via an agentic approach}
\label{fig:overall-framework}
\end{figure}

To build an autonomous disruption monitoring framework, we require four key stages, as illustrated in \autoref{fig:overall-framework}: (1) \textit{Event Detection \& Data Ingestion} (2) \textit{Relevance Filtering} (3) \textit{Supply Chain Risk Assessment} and (4) \textit{Action Planning and Execution}. These stages operate sequentially: each stage uses outputs from the previous stage as inputs. Each stage has three defining characteristics: a distinct purpose, specific input data requirements, and dedicated agents for execution. In \textbf{Stage~1}, potential disruptions are detected from external sources such as news feeds, public alerts, and API-based event streams. In \textbf{Stage~2}, these detected disruptions are evaluated for their relevance and significance to the monitored company, filtering out events that are not operationally impactful. In \textbf{Stage~3}, the validated disruptions are traced through the company's extended supply chain to determine whether the affected entities exist within its direct or indirect network. This stage identifies disruption paths and quantifies the corresponding risk exposure. Finally, in \textbf{Stage~4}, the framework formulates data-driven mitigation strategies based on the nature and severity of the disruption, as well as its assessed risk. These may include recommending alternative suppliers, adjusting sourcing routes, or proposing actions to minimise cascading impacts across the supply chain.

In our system, an \textit{agent} is represented as a language model-based autonomous unit designed to perform a specific reasoning task within the supply chain disruption monitoring framework. {Each agent is configured through a carefully engineered prompt that defines its (1) role, (2) task boundaries, and (3) output format.} {This design enables agents to operate independently as they reason over complex supply chain information, and interact with external tools and other agents in the system.}

\begin{figure}[H]
\centering
    \includegraphics[width=1\linewidth]{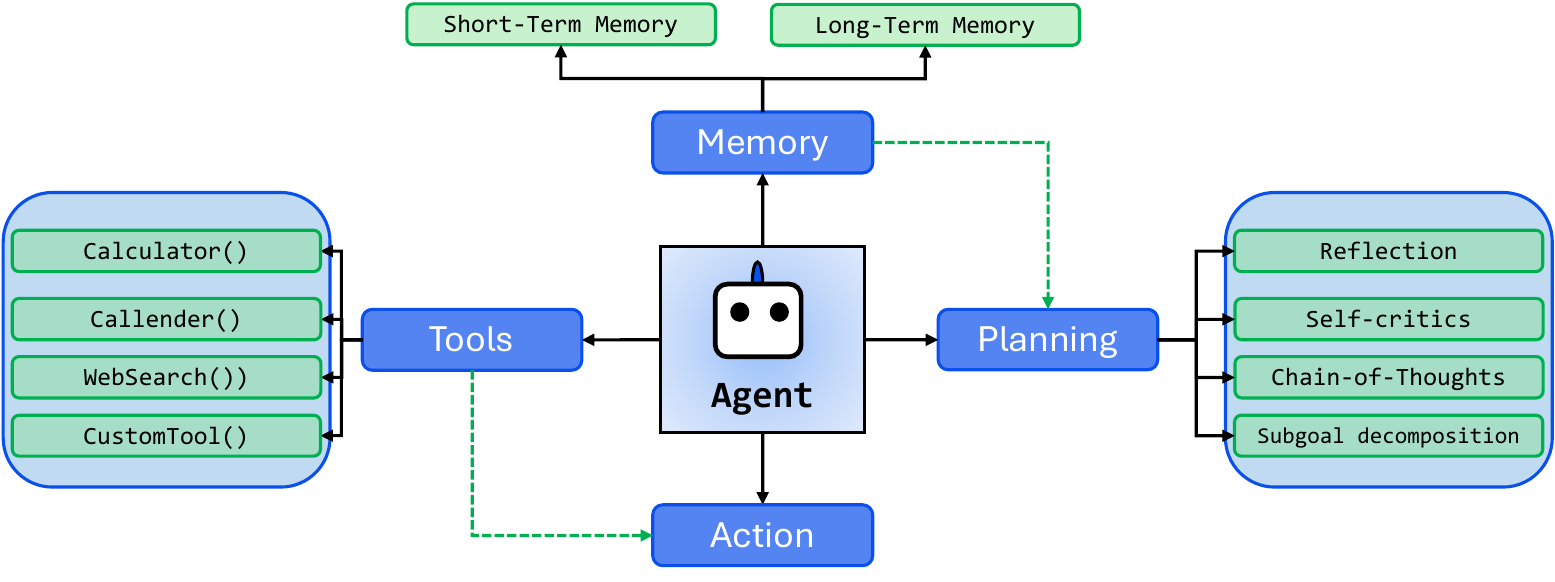}
    \caption{Modular architecture of an autonomous agent. Each agent is composed of four key components: tools, memory, planning, and actions, which together enable it to reason, retrieve relevant information, and generate structured outputs.}
\label{fig:single-agent}
\end{figure}

As shown in \autoref{fig:single-agent}, each agent is structured around four core components: 
(a) \textit{tools}, which allow it to retrieve or manipulate data; 
(b) \textit{memory}, which supports context retention across steps; 
(c) \textit{planning}, which enables task decomposition and reasoning; and 
(d) \textit{actions}, which define its outputs or downstream communications. {This modular structure draws on established principles from agentic systems literature \citep{wooldridge1995intelligent, jennings1993commitments}, and integrates recent advancements in large language model orchestration.}

\begin{figure}[H]
\centering
    \includegraphics[width=\linewidth]{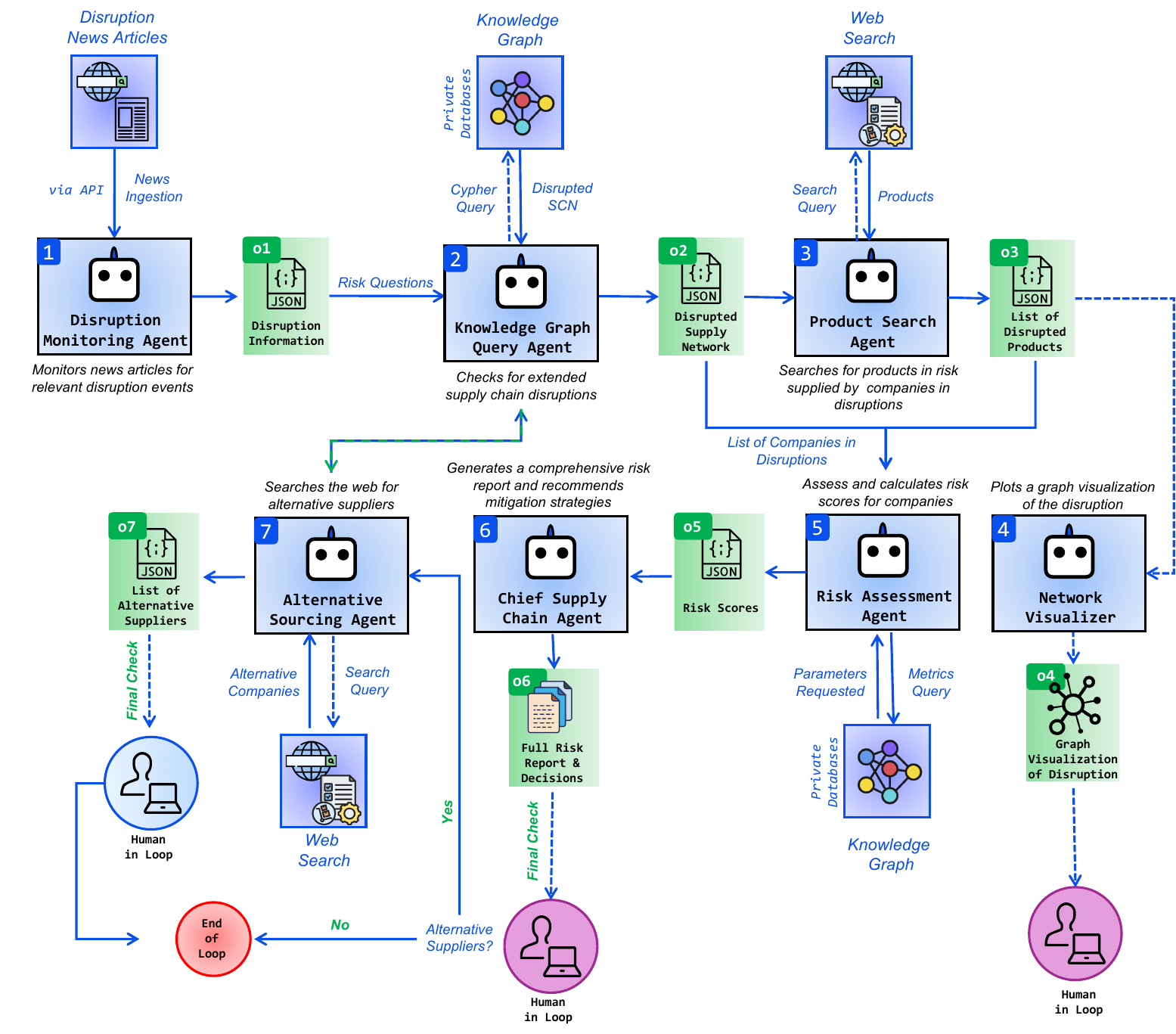}
    \caption{Proposed framework for autonomous agentic disruption monitoring. The framework consists of seven specialised agents (numbered blue boxes), each performing a distinct reasoning task in the end-to-end supply chain risk assessment pipeline. Outputs of each agent are shown in green and denoted as \texttt{o-\#} for clarity and traceability across stages.}
\label{fig:framework}
\end{figure}

To implement these four stages, we deploy seven specialised agents that collectively execute the stage objectives. {Each stage may include one or more agents working together to achieve its goals.}  \autoref{fig:framework} presents the architecture of our agentic disruption monitoring framework.  {Each agent is a specialized agent assigned a distinct task relevant to one of the stages in the end-to-end disruption analysis process.} {These agents operate sequentially: the output of one agent becomes the structured input for the next.} {Each output is serialized in JSON\footnote{JSON (JavaScript Object Notation) is a lightweight data interchange format that uses human-readable text to represent structured data objects.} to ensure consistency and interoperability.} {Each agent is provided with dedicated tools aligned with its functional responsibilities.} {These include knowledge graph querying utilities, custom metric calculators for risk scoring, and external web search tools.} {Tools are embedded directly into the agent's reasoning flow, enabling autonomous data retrieval and analysis.} {A detailed overview of agents, their roles, tasks, and tools is provided in \autoref{sec:agent-decomposition}.}

\rev{The architecture presented here is one possible configuration for implementing the four-stage disruption monitoring objectives. {We organise agents by role and task specialisation, where each agent is assigned a distinct responsibility aligned with a specific stage objective.} {This role-based decomposition enables clear separation of concerns, facilitates independent development and testing of each agent, and supports modular adaptation to different operational contexts.} {Alternative architectures are possible, such as combining multiple tasks into fewer agents or further decomposing tasks into additional specialised agents, as long as the four-stage objectives are achieved accurately.}} {In our implementation, agents are developed using CrewAI agent framework \citep{CrewAI:Platform}, which manages scheduling, memory, tool integration, and inter-agent communication.} {All agents use OpenAI's GPT-4o language model \citep{hurst2024gpt}, selected for its strong performance in structured reasoning and reliable schema-conforming outputs.} {The system architecture is model- and framework-agnostic and can be adapted to other agent development platforms or LLMs as needed.}

\subsection{Agent Decomposition: Roles, Tasks \& Tools}
\label{sec:agent-decomposition}

In this section, we decompose and discuss the role, tasks, and tools for each of the seven agents used in our framework. {\autoref{tab:agent_summary} provides a summary of each agent's role, inputs, and outputs.} {\autoref{tab:tool_summary} lists all tools provided to agents to execute their tasks.} {In the following subsections, we discuss the design of each agent's prompt, the tools it invokes, and the schema of its inputs and outputs.} {As this framework is designed for a supply chain management application, we make sure to emphasise the supply chain management implications of each agent's functionality.}

{Before deploying this framework, organisations must have access to an extended supply chain network of the company in interest.} {This is a prerequisite for framework operation.} {The network must include multi-tier extended supplier relationships (typically up to Tier-4), company locations, and industry classifications.} {Such networks can be obtained from private data providers such as Bloomberg\footnote{\url{https://www.bloomberg.com/professional/solutions/corporations/supply-chain/}}, FactSet\footnote{\url{https://www.factset.com}}, or Panjiva\footnote{\url{https://panjiva.com/}}, or constructed from internal supplier databases and procurement records.} {The framework's analytical capabilities depend entirely on this network data.} {Without it, agents cannot traverse supplier paths, identify cascading risks, or quantify exposure metrics.} {The network structure is the foundation upon which all framework operations are built.}

\begin{table}[h]
  \centering
\caption{Overview of agent responsibilities and data interfaces in the disruption monitoring pipeline.}
  \label{tab:agent_summary}
\resizebox{\textwidth}{!}{%
    \begin{tabular}{@{}l
        >{\raggedright\arraybackslash}p{5.5cm}
        >{\raggedright\arraybackslash}p{4cm}
        >{\raggedright\arraybackslash}p{4cm}@{}}
      \toprule
      \textbf{Agent} & \textbf{Function} & \textbf{Input} & \textbf{Output} \\
      \midrule
      Disruption Monitoring Agent       
        & Detects and classifies disruption events in unstructured text; formulates diagnostic queries      
        & News articles (RSS, scraped HTML, social media posts)      
  & JSON: disruption type, summary, identified entities, follow-up questions \\
      \addlinespace
      Knowledge Graph Query Agent 
  & Resolves entity names; traces multi-tier supplier paths via Cypher traversals (Neo4j's graph query language)  
        & JSON: disruption entities and diagnostic queries     
  & JSON: tier-annotated supplier chains exposed to disruption\\
      \addlinespace
      Product Search Agent      
        & Annotates each supplier link with specific products or materials supplied      
        & JSON: supplier chains from KG Query      
        & JSON: company--product mappings preserving tier order \\
      \addlinespace
      Risk Manager Agent        
        & Computes Tier-1 supplier risk scores by aggregating exposure depth, exposure breadth, downstream criticality, and network centrality metrics      
        & JSON: disrupted supply chain paths from KG Query      
        & JSON: risk scores, risk levels, and component breakdowns for top 10 riskiest Tier-1 suppliers \\
      \addlinespace
      Network Visualizer Agent   
        & Generates a tiered network diagram of the network in disruption      
        & JSON: subgraph and risk metrics      
        & HTML render of annotated disrupted network \\
      \addlinespace
      CSCO Agent  
        & Synthesizes analytics into an expert action plan      
  & JSON: risk assessment results, disrupted product--company chain      
        & Structured report: mitigation recommendations \\
      \addlinespace
      Alternative Sourcing Agent    
  & Identifies and validates substitute suppliers for high-risk components      
  & JSON: action items from supply-chain agent for alternative supplier sourcing      
        & JSON: identified alternative suppliers for the same product in risk \\
      \bottomrule
\end{tabular}%
}
\end{table}

\begin{table}[h]
  \centering
\caption{Catalogue of tools used across agents in the disruption monitoring framework.}
  \label{tab:tool_summary}
\resizebox{\textwidth}{!}{%
    \begin{tabular}{@{}l
        >{\raggedright\arraybackslash}p{6cm}
        >{\raggedright\arraybackslash}p{3.7cm}
        >{\raggedright\arraybackslash}p{3.1cm}@{}}
      \toprule
      \textbf{Tool} & \textbf{Function} & \textbf{Used By} & \textbf{Type} \\
      \midrule
      \texttt{ScrapeWebsite()} 
        & Extracts raw textual content from URLs or API endpoints
        & Disruption Monitoring Agent
  & Built-In (CrewAI)\\
      \addlinespace
      \texttt{resolve\_entity\_struct} 
  & Standardizes free-text company names into canonical identifiers used in the knowledge graph
        & Knowledge Graph Query Agent 
        & Custom \\
      \addlinespace
      \texttt{company\_list\_struct} 
  & Generates tier-specific lists of companies to support query composition
        & Knowledge Graph Query Agent 
        & Custom \\
      \addlinespace
      \texttt{calculate\_tier\_struct} 
  & Annotates nodes with tier-depth relative to the focal firm
        & Knowledge Graph Query Agent 
        & Custom \\
      \addlinespace
      \texttt{supply\_chain\_bfs\_struct} 
  & Constructs breadth-first queries for traversing supplier chains up to Tier-4
        & Knowledge Graph Query Agent 
        & Custom \\
      \addlinespace
      \texttt{SerperDevTool} 
        & Performs web searches to extract product catalogues or identify alternative suppliers
        & Product Search Agent, Alternative Sourcing Agent 
        & External API \\
      \addlinespace
      \texttt{tier1\_comprehensive\_risk\_tool} 
        & Computes Tier-1 supplier risk scores by aggregating exposure depth, exposure breadth, downstream criticality, and network centrality metrics
        & Risk Manager Agent 
        & Custom \\
      \addlinespace
      \texttt{networkx\_plot\_struct} 
        & Generates tiered visualizations of disrupted supply networks with risk annotations
        & Network Visualizer Agent 
        & Custom \\
      \bottomrule
\end{tabular}%
}
\end{table}

\begin{figure}[H]
\centering
    \includegraphics[width=1\linewidth]{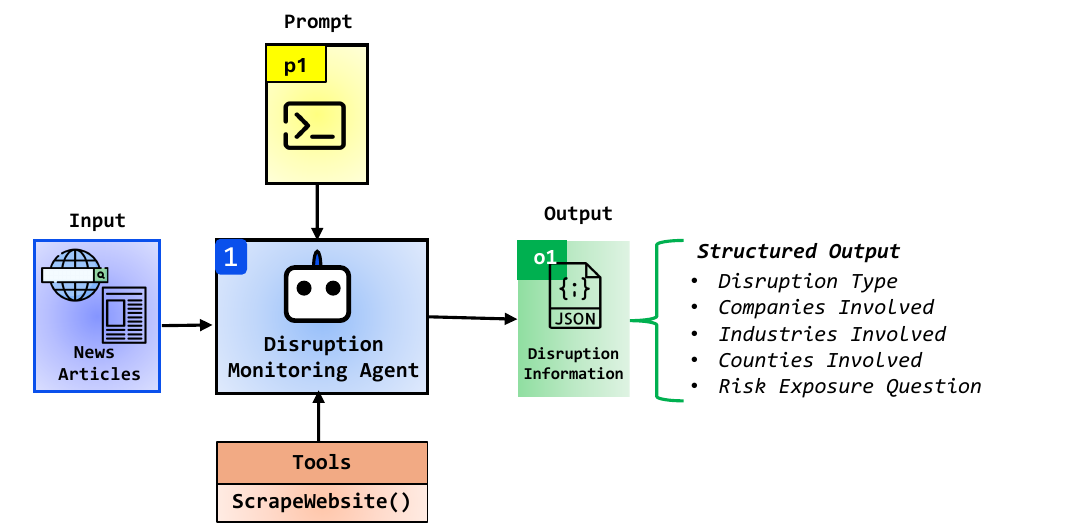}
    \caption{Agent 1: Disruption Monitoring Agent - Input, output, tools and prompt}
\label{fig:agent1}
\end{figure}

\subsubsection{Disruption Monitoring Agent}

\rev{The Disruption Monitoring Agent (Agent 1 in \autoref{fig:framework}) serves as the entry point of the disruption analysis pipeline, determining whether a supply chain disruption has occurred by examining news articles from credible sources.} \rev{The agent adopts the reasoning perspective of an experienced senior supply chain risk analyst, employing chain-of-thought prompting \citep{TomHardware2022} to analyse article content and classify disruption types (geopolitical, economic crisis, natural disaster, cybersecurity incident).} \rev{It extracts critical entities including affected countries, industries, and companies, and formulates structured diagnostic questions that trace the monitored company's extended supplier network up to Tier-4.} \rev{These questions are designed to be executable against the knowledge graph, enabling systematic identification of suppliers potentially affected by the disruption.} \rev{The agent's output, structured as JSON, includes disruption classification, identified entities, and diagnostic queries that guide subsequent network traversal.} \rev{This agent transforms unstructured news information into actionable intelligence that enables systematic investigation of potential supply chain impacts (See \autoref{fig:agent1})}

\begin{figure}[H]
\centering
    \includegraphics[width=1\linewidth]{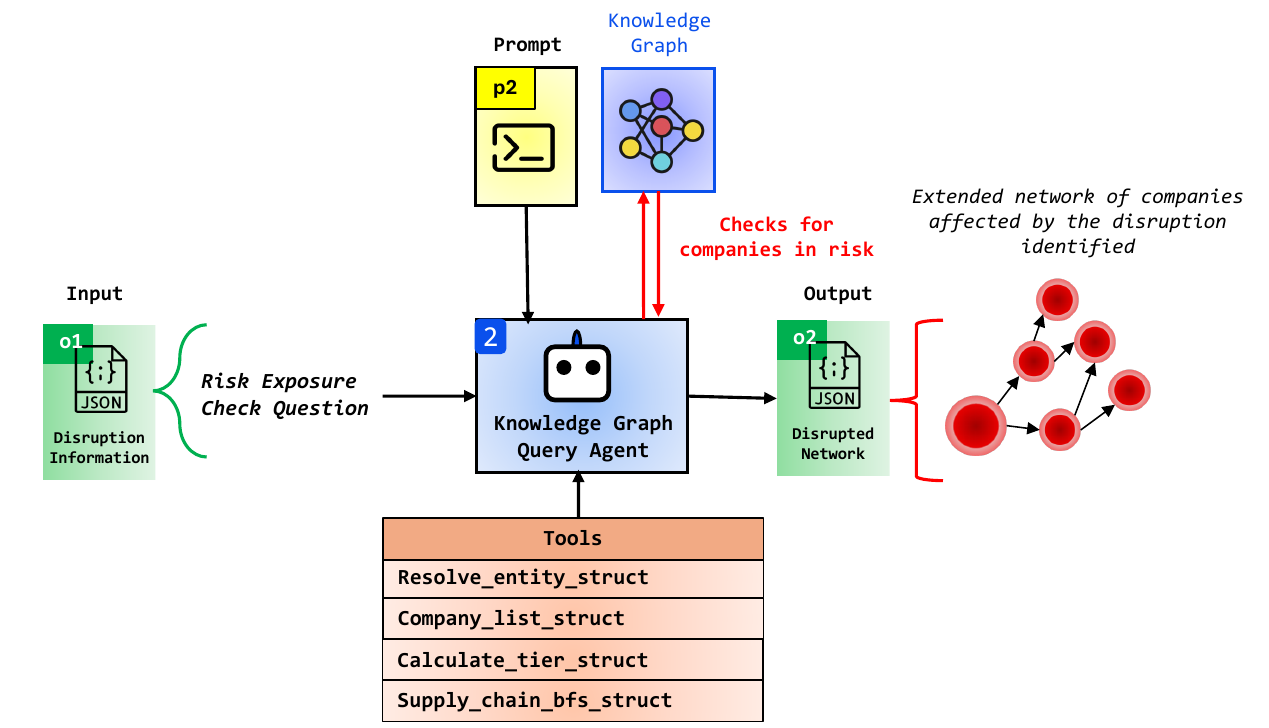}
    \caption{Agent 2: Knowledge Graph Query Agent architecture, tools, inputs used and output structure}
\label{fig:agent2}
\end{figure}

\subsubsection{Knowledge Graph Query Agent} \label{section:kg-query-agent}

\rev{The Knowledge Graph Query Agent (Agent 2 in \autoref{fig:framework}) translates structured diagnostic questions from the Disruption Monitoring Agent into precise graph database queries, systematically mapping multi-tier supplier relationships affected by identified disruptions.} \rev{The agent operates over a structured supply chain knowledge graph stored in Neo4j\footnote{Neo4j is a graph database management system that stores data as nodes and relationships, enabling efficient traversal of complex network structures.} \citep{Neo4j2024Neo4j:Platform}, where nodes represent companies, countries, and industries, and edges represent supplier relationships (\texttt{suppliesTo}) and geographic attributes (\texttt{locatedIn}) \citep{almahri2024enhancing}.} \rev{The agent employs entity resolution to standardise company identifiers and constructs complete supplier dependency paths extending from the focal firm down to Tier-4 suppliers using breadth-first traversal.} \rev{Queries are automatically filtered based on disruption criteria (country, industry, tier) to return only relevant supply chain relationships.} \rev{The agent translates natural language questions into structured Cypher\footnote{Cypher is a declarative graph query language used by Neo4j to traverse and query graph databases efficiently.} queries, enabling systematic network traversal without requiring prior knowledge of graph database query languages.} \rev{The output is a fully expanded, tier-resolved JSON structure containing all supplier paths that intersect with the disruption source, with metadata for company name, country, and industry as shown in \autoref{fig:agent2}}

\begin{figure}[H]
\centering
    \includegraphics[width=1\linewidth]{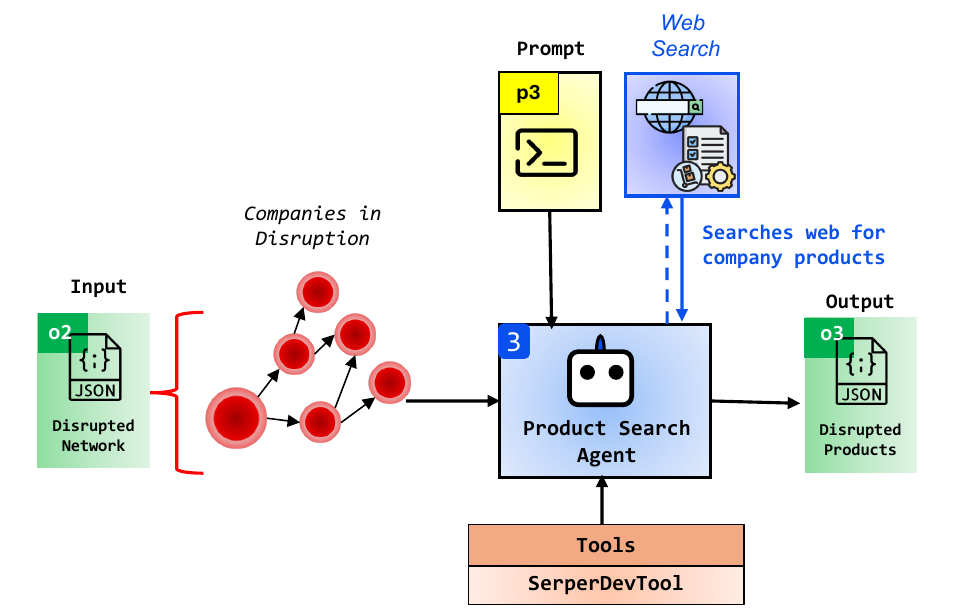}
    \caption{Agent 3: Product Search Agent architecture, tools, inputs and outputs structure}
\label{fig:agent3}
\end{figure}

\subsubsection{Product Search Agent}

\rev{The Product Search Agent (Agent 3 in \autoref{fig:framework}) enriches supplier chains with product-level information, determining what materials, components, or services are supplied at each link in the disrupted supply chain.} \rev{The agent systematically processes every supplier relationship identified by the Knowledge Graph Query Agent, leveraging web search capabilities to infer precise products being transferred between suppliers.} \rev{The output is a structured JSON object mirroring the original supply paths, with an additional field specifying the product supplied at each relationship.} \rev{This product-level annotation enables downstream agents to assess which specific components may be impacted by a disruption, supporting targeted risk scoring and mitigation planning (\autoref{fig:agent3}).} 

\subsubsection{Network Visualizer Agent}

\rev{The Network Visualizer Agent (Agent 4 in \autoref{fig:framework}) translates the annotated supply chain graph into a visual diagram to support interpretation and decision-making.} \rev{The agent receives tier-structured supplier chains and computed risk metrics as input, generating a network plot where node size and colour reflect risk level and importance, with high-risk nodes labeled in red and edges labeled with associated product flows.} \rev{The output is rendered as an interactive HTML view, enabling decision-makers to visually assess disruptions, quickly identify critical suppliers, and understand disruption propagation patterns across the network.}

\subsubsection{Risk Manager Agent}

\rev{The Risk Manager Agent (Agent 5 in \autoref{fig:framework}) quantifies risk exposure for Tier-1 suppliers by aggregating downstream disruption impacts from deeper tiers.} \rev{The agent focuses on Tier-1 suppliers because companies typically have direct operational control only over Tier-1 suppliers, yet disruptions in deeper tiers can cascade to affect Tier-1 suppliers and ultimately the monitored company.} \rev{The agent computes comprehensive risk scores using deterministic mathematical calculations that aggregate four risk dimensions: exposure depth (maximum disrupted tier), exposure breadth (count of disrupted downstream suppliers weighted by tier proximity), downstream criticality (aggregated criticality of disrupted downstream suppliers), and network centrality (graph-based centrality measures).} \rev{The risk score is computed as a weighted combination: 35\% exposure breadth, 25\% dependency ratio, 20\% downstream criticality (maximum of downstream centrality or PageRank), 10\% Tier-1 supplier centrality, and 10\% exposure depth (normalized by maximum tier).} \rev{These calculations are deterministic, ensuring reproducibility given the same disrupted paths.} \rev{The agent filters to the top 10 riskiest suppliers and structures the output as JSON, including risk scores, risk levels (HIGH, MEDIUM, LOW based on thresholds of 0.6, 0.45-0.59, and $<$0.45), and component breakdowns showing the contribution of each risk dimension.} This agent enables decision-makers to focus on suppliers where they have operational control while accounting for risks originating in deeper tiers, providing actionable intelligence for strategic decision-making through the aggregation of cascading effects into Tier-1 risk scores.

\subsubsection{Chief Supply Chain Officer (CSCO) Agent}

\rev{The CSCO Agent (Agent 6 in \autoref{fig:framework}) functions as the central decision-making unit, synthesizing all upstream outputs, disruption details, supplier paths, and risk metrics into a coherent response plan.} \rev{The agent assumes the role of a Chief Supply Chain Officer, generating targeted mitigation strategies that encompass actionable recommendations: supplier replacement, increased monitoring, or maintaining standard operations.} \rev{The agent maps risk scores to actions using deterministic thresholds: HIGH risk suppliers (risk score $\geq$ 0.6) should be replaced, MEDIUM risk suppliers (risk score 0.45-0.59) should have increased monitoring, and LOW risk suppliers (risk score $<$ 0.45) should maintain standard operations.} \rev{The agent also generates qualitative analysis sections including disruption summary, network impact analysis, and replacement recommendations, providing executive-level communication that supports decision-making with clear reasoning.} \rev{The action plan is routed to a human reviewer for approval, revision, or override, with approved plans proceeding to the Alternative Sourcing Agent where applicable.} \rev{From a supply chain management perspective, this agent enables decision-makers to prioritise actions based on risk exposure, providing both quantitative decisions and qualitative justifications that address the need for executive-level communication where decisions must be supported by clear reasoning.}

\subsubsection{Alternative Sourcing Agent}

\rev{The Alternative Sourcing Agent (Agent 7 in \autoref{fig:framework}) operates as the system's procurement specialist, searching for replacement suppliers for high-risk or disrupted components upon receiving the executive decision plan from the CSCO Agent.} \rev{The agent queries external web sources to identify alternative suppliers for specific products, then interacts with the Knowledge Graph Query Agent to verify that their extended multi-tier supply chains are free from the previously identified disruption.} \rev{The verified alternative supplier data is structured into a JSON object and routed to human reviewers for final validation.} \rev{This agent enables rapid identification of alternative suppliers when disruptions require supplier replacement, supporting business continuity and operational resilience.} \rev{The framework can be extended with multi-criteria decision analysis (MCDA) to rank alternatives using criteria such as cost, lead time, quality, capacity, geographic risk, and ESG performance \citep{ho2010multi,chai2020decision,govindan2015multi}, and with dynamic Bayesian networks (DBN) to model disruption propagation over time and across tiers \citep{sharma2022supply, liu2021new}.}

\subsection{Prompt Engineering Design Principles}

Each agent in the agentic system is guided by a carefully designed prompt that defines its role, context, and expected output format. {We use role-based instructions via system messages to specify each agent's persona and responsibilities, following best practices in LLM deployment \citep{zamfirescu2023johnny, wei2022chain}.} {Prompt templates break down complex tasks into subtasks or sequential steps (multi-step decomposition).} {They explicitly ask the model to ``think step by step,'' leveraging chain-of-thought prompting to elicit intermediate reasoning.} {Some prompts include exemplar dialogues or sub-problem breakdowns that guide the LLM's reasoning process \citep{wei2022chain, reynolds2021prompt}.}

{We employ few-shot prompting, providing illustrative examples of inputs paired with structured outputs.} {This demonstrates the desired format and reasoning style, increasing output quality \citep{Brown2020LanguageLearners}.} {All prompts enforce a strict output schema: agents are instructed to output JSON records with predefined fields.} {This ensures consistency and machine-parseability of responses.} {We incorporate explicit format instructions (e.g., ``Respond with a JSON object having these keys...'').} {This design was informed by recent work on structured output.} {GPT-4o's ``Structured Outputs'' feature guarantees that model responses exactly match a provided JSON schema.}

{In our system, each agent's prompt concludes with a template defining the output structure (see \autoref{full-prompts} for full prompt templates).} {By anchoring the LLM with role-based context and schema guidance, we obtain predictable, structured outputs across all agents.} {Prompt design plays a critical role in shaping agent behaviour.} {Even minor modifications to the prompt can significantly alter the agent's reasoning and output quality.} {Strict output schemas ensure that agent outputs can be reliably integrated into operational decision-making processes, where consistency and predictability are essential.}

\subsection{Agent Communication and Execution Flow}

Agents exchange information in a structured, serialized form to avoid ambiguity and facilitate parsing. {In our implementation, inter-agent messages are carried in JSON formats.} {This follows common practice in agentic architectures: modern systems typically communicate across agents through JSON payloads delivered via API calls \citep{dahling2021enabling}.} {Using JSON (rather than free-text) ensures that each field is unambiguous and machine-readable.} {JSON is widely supported and aligns with LLM structured-output features.}

{The execution model is modular and sequential.} {Each agent operates as a distinct module in a directed workflow graph.}  {That agent processes the input and emits its result in the same serialized format, which is then passed to the next agent in the sequence.} {The system behaves like a stateful pipeline: at each step, the agent node receives the current state, executes its task, and passes the updated state to the next nodes.}

{This modular chain-of-command aligns with established agentic design: separating concerns into specialised agents improves maintainability and scalability.} {Each agent only needs to handle its own subtask and output schema, simplifying development and testing.} {From a supply chain management perspective, this modular design enables organisations to adapt individual agents to their specific operational needs while maintaining the overall system structure.}

\subsection{System Configuration and Modularity Considerations}

This framework is designed as a flexible template for implementing supply chain monitoring with LLM-powered agents. {In our reference setup, we used OpenAI's GPT-4o model for all agents due to its strong performance in structured reasoning and reliability in returning valid JSON outputs.} {However, the framework is model-agnostic and can be adapted to other API-accessible LLMs such as GPT-4, Claude~3, or Mistral, or locally deployed models, depending on performance, cost, or availability.}

{The knowledge graph used for querying is assumed to exist beforehand.} {In our case, it is implemented using Neo4j and encodes companies, countries, industries, and supplier relationships based on the schema described in \cite{almahri2024enhancing}.} {Users may substitute their own graph databases or structured tables, as long as they adjust the prompts and agent logic to match their data format and ontology.}

{News articles used for disruption monitoring may come from any reliable data source.} {Users set up their own ingestion pipeline using trusted RSS feeds, commercial news APIs, or internal monitoring systems.} {It is essential to include a basic filtering step to ensure that only relevant and credible reports are passed to the agent.}

{Product lookup and supplier search functions depend on external tools.} {We use the Serper.dev API to retrieve public product or supplier information via web search.} {This can be replaced or augmented with internal databases if preferred.} {Each tool is modular and can be swapped out as needed without changing the overall structure of the system.}

{Prompt templates for each agent are modular and can be adjusted for different industries, data formats, or reasoning strategies.} {Few-shot examples and system messages can be tuned to reflect the specific domain or operational style of the user.} {Users can extend the output schemas to include additional metadata such as delivery lead times, ESG ratings, or supplier certifications.}

{To help users judge reliability, adding a per-agent confidence score that reflects the agent's certainty in its own task output could be useful.} {This can be a simple 0--100 score or Low/Medium/High.} {Prior work shows language models can report their own confidence and that these reports can be calibrated \citep{kadavath2022language,tian2023just,geng2023survey}.} {We do not include confidence outputs in this manuscript; we recommend them for practical deployments.}

{Overall, the system is adaptable.} {Users can integrate their own data, tools, and models, provided that agents communicate through clearly defined schemas and follow a structured task decomposition.} {\autoref{full-prompts} includes all prompt templates and tool definitions used in our implementation.} {From a supply chain management perspective, this modularity enables organisations to customise the framework to their specific operational contexts, data sources, and risk tolerance levels, while maintaining the core analytical capabilities.}


\rev{\section{Experimental Evaluation}
\label{sec:evaluation}

\subsection{Experimental Design and Methodology}

This section presents an experimental evaluation of our agentic supply chain disruption monitoring framework. The evaluation assesses four capabilities: (1) extracting disruption information from unstructured news articles, (2) identifying disrupted supply chain paths across multiple tiers, (3) quantifying risk exposure for Tier-1 suppliers, and (4) generating strategic decisions aligned with disruption profiles. We evaluate the framework across three automotive manufacturers (Tesla Inc., Mercedes-Benz Group AG, and Bayerische Motoren Werke AG) using 30 synthesized disruption scenarios (10 per company). These scenarios capture diverse disruption types, geographic contexts, and supply chain depths. The scenario synthesis methodology is explained in more detail in \autoref{scenario-synthesis}.

\subsection{Dataset and Evaluation Methodology}

\subsubsection{Knowledge Graph Dataset}

Our evaluation uses a real-world supply chain network stored in a Neo4j knowledge graph. The network is extracted from global supply chain data described in our previous work \citep{almahri2024enhancing}. The dataset contains 6,596 nodes and 23,888 relationships, forming a large-scale supply chain network centered on electric vehicle manufacturing. The dataset includes 6,495 company nodes distributed across 101 country nodes and 25 distinct industries (stored as properties on Company nodes). The dataset contains 17,341 supply chain relationships (\texttt{suppliesTo}) connecting suppliers to customers across multiple tiers, and 6,547 location relationships (\texttt{locatedIn}) linking companies to their geographic locations.

For this evaluation, we focus on three major automotive manufacturers whose supply chains are embedded within this larger network. Table~\ref{tab:company_supply_chains} presents the supply chain characteristics for each monitored company, extracted by traversing \texttt{suppliesTo} relationships from the company node (Tier-0) through Tier-4 suppliers. These firms were selected because they offer variation in supplier depth, geographic coverage, and network complexity. These multi-tier networks allow us to test disruption propagation and framework performance in realistic conditions.

\begin{table}[h]
\centering
\caption{Supply Chain Network Characteristics for Monitored Companies}
\label{tab:company_supply_chains}
\resizebox{0.7\textwidth}{!}{%
\begin{tabular}{lcccc}
\toprule
\textbf{Company} & \textbf{Tier-1} & \textbf{Tier-2} & \textbf{Tier-3} & \textbf{Tier-4} \\
\midrule
Tesla Inc & 34 & 242 & 721 & 1,042 \\
Mercedes-Benz Group AG & 33 & 88 & 978 & 2,467 \\
Bayerische Motoren Werke AG & 19 & 97 & 603 & 2,699 \\
\bottomrule
\end{tabular}%
}
\end{table}

\subsubsection{Scenario Synthesis} \label{scenario-synthesis}

Due to the lack of publicly available supply chain disruption datasets with extended disrupted network structures, we manually synthesized disruption scenarios based on the knowledge graph dataset for the three selected companies. Our evaluation scenarios were manually synthesized by two domain experts  working within supply chain risk management, through systematic analysis of each company's network structure. This manual process ensures that scenarios are strategically designed to test specific capabilities and edge cases, rather than randomly generated patterns that may miss critical failure modes.

We synthesized 30 evaluation scenarios (10 per company) covering five distinct disruption types: economic crises (15 scenarios), geopolitical events (6 scenarios), labour strikes (3 scenarios), natural disasters (3 scenarios), and cybersecurity incidents (3 scenarios). These categories reflect high-impact, real-world disruptions affecting supply chains. The 30 scenarios are distributed as 23 true positive scenarios, where disrupted supply chain paths exist in the knowledge graph, and 7 false positive scenarios, where no disrupted paths exist. This distribution tests the framework's ability to correctly reject irrelevant disruptions and ensures comprehensive evaluation of both detection accuracy and precision.

Furthermore, the scenarios synthesized target different network tiers. This enables a targeted assessment of the framework's ability to handle disruptions at various depths: Tier-1 (3 scenarios) for direct supplier disruption, testing immediate impact assessment, Tier-2 (9 scenarios) for second-tier supplier disruptions, testing multi-tier propagation analysis, Tier-3 (3 scenarios) for third-tier supplier disruptions, testing deep supply chain traversal, and Tier-4 (15 scenarios) for fourth-tier supplier disruptions, testing the framework's ability to handle the deepest and most complex multi-tier disruptions. The emphasis on Tier-4 scenarios (50\% of all scenarios) reflects the critical importance of detecting deep supply chain disruptions, where impacts may be less immediately apparent but can cascade through multiple tiers to affect the monitored company.

\subsubsection{Ground Truth Dataset Generation}

Domain experts manually performed all tasks that the agents are designed to execute, generating a comprehensive ground truth dataset to benchmark agent outputs. Experts independently generated labeled datasets to prevent any bias. Any disagreements were resolved through revaluation. {We focus our evaluation on four core agents that constitute the essential part of the framework: the Disruption Monitoring Agent, the Knowledge Graph Query Agent, the Risk Manager Agent, and the CSCO Agent.} {These four agents were selected for evaluation because they represent the core analytical pipeline with objectively quantifiable outputs that can be rigorously compared against manually generated ground truth.} {The remaining three agents (Product Search, Visualization, and Alternative Sourcing) serve supplementary functions that are either optional enhancements, require subjective quality assessment, or depend on external data sources not available in our evaluation framework.}

For the Disruption Monitoring Agent, experts manually extracted the disruption type, affected countries, industries, and companies explicitly mentioned in each scenario's news article text. {This ensures that only entities directly referenced in the source material were included.} {For the KG Query Agent, experts manually traced all disrupted supply chain paths from each monitored company through Tier-4 suppliers using the Neo4j knowledge graph.} {They followed supplier-to-customer relationships and identified paths containing companies located in affected countries or operating in affected industries.} {Complete chain structures were captured including company names, countries, industries, and tier classifications.} {For the Risk Manager Agent, experts manually calculated risk scores for each Tier-1 supplier by aggregating four quantitative metrics: exposure depth (maximum disrupted tier), exposure breadth (count of disrupted downstream suppliers weighted by tier proximity), downstream criticality (aggregated criticality of disrupted downstream suppliers), and network centrality (graph-based centrality measures).} {The risk score is computed as a weighted combination: 35\% exposure breadth, 25\% dependency ratio, 20\% downstream criticality, 10\% Tier-1 supplier centrality, and 10\% exposure depth.} {For the CSCO Agent, experts manually determined expected decisions for the top 10 riskiest Tier-1 suppliers based on deterministic rules: HIGH risk suppliers (risk score $\geq$ 0.6) should be replaced, MEDIUM risk suppliers (risk score 0.45-0.59) should have increased monitoring, and LOW risk suppliers (risk score $<$ 0.45) should maintain standard operations.} {These thresholds are configurable parameters and can be adjusted based on the user's operational risk tolerance, strategic priorities, or industry-specific risk appetite.} {This manually generated ground truth dataset serves as the benchmark for evaluating agent performance, ensuring that the evaluation reflects expert judgment and comprehensive analysis.}

\subsubsection{Evaluation Metrics}

We employ standard information retrieval metrics to evaluate agent performance, following established practices in information extraction, knowledge graph querying, and agentic system evaluation. The fundamental metrics are defined as:

\begin{align}
\text{Precision} &= \frac{\text{TP}}{\text{TP} + \text{FP}} \\
\text{Recall} &= \frac{\text{TP}}{\text{TP} + \text{FN}} \\
\text{F1 Score} &= \frac{2 \times \text{Precision} \times \text{Recall}}{\text{Precision} + \text{Recall}}
\end{align}

where TP, FP, and FN denote true positives, false positives, and false negatives, respectively. Macro-averaged metrics are computed as the mean and standard deviation across scenarios: $\bar{M} = \frac{1}{N}\sum_{i=1}^{N} M_i$ and $\sigma_M = \sqrt{\frac{1}{N-1}\sum_{i=1}^{N}(M_i - \bar{M})^2}$, where $M_i$ represents the metric value for scenario $i$ and $N$ is the number of scenarios.

For the Disruption Monitoring Agent, we evaluate entity extraction across countries, industries, and companies. A true positive occurs when an entity appears in both agent output and ground truth. A false positive occurs when an entity appears only in agent output or when extraction is partially correct but incomplete. A false negative occurs when an entity appears only in ground truth. Metrics are averaged across the three entity types.

For the Knowledge Graph Query Agent, we evaluate the framework's ability to identify disrupted supply chain paths across all four tiers (Tier-1 through Tier-4). Each path is represented as a set of (company, country, industry) nodes. Paths are compared using order-insensitive set matching. We employ Jaccard similarity with a threshold of 0.9 to handle partial matches, where paths with Jaccard similarity $\geq$ 0.9 are considered matches. This approach accounts for minor variations in path representation while maintaining rigorous evaluation standards. A true positive occurs when a path appears in both outputs. A false positive occurs when a path appears only in agent output or is incomplete. A false negative occurs when a path appears only in ground truth. Metrics are averaged across all tiers.

For the Risk Manager Agent, we evaluate supplier identification and risk score accuracy for the top 10 riskiest Tier-1 suppliers. A supplier is correctly identified (TP) if the name matches the ground truth and the risk score matches within 0.1 tolerance. This tolerance accounts for floating-point variations in graph algorithm computations. False positives occur when suppliers are incorrectly identified or scores differ by more than 0.1. False negatives occur when suppliers are missed.

For the CSCO Agent, we employ both quantitative and qualitative evaluation. {The quantitative evaluation uses Precision, Recall, and F1 Score, where the agent must address all top 10 riskiest suppliers with appropriate actions aligned with risk levels.} {A true positive occurs when a supplier is correctly addressed.} {A false positive occurs when addressed with an incorrect action.} {A false negative occurs when not addressed.} {The qualitative evaluation employs a rubric-based methodology, assessing disruption summary, network impact analysis, and replacement recommendations across weighted criteria (completeness, clarity, accuracy, relevance, insightfulness, and actionability), as detailed in \autoref{subsec:csco_qualitative}.}

\subsection{Results and Discussion}

Table~\ref{tab:overall_metrics} presents the overall performance metrics aggregated across all evaluation scenarios using macro-averaging. All scenarios completed successfully.

\begin{table}[h]
\centering
\caption{Overall Performance Metrics (Mean $\pm$ Std) Tested over 30 Scenarios}
\label{tab:overall_metrics}
\resizebox{0.7\textwidth}{!}{%
\begin{tabular}{lccc}
\toprule
\textbf{Agent} & \textbf{Precision} & \textbf{Recall} & \textbf{F1 Score} \\
\midrule
Disruption Monitoring & $0.983 \pm 0.051$ & $1.000 \pm 0.000$ & $0.991 \pm 0.028$ \\
KG Query              & $1.000 \pm 0.000$ & $0.975 \pm 0.137$ & $0.980 \pm 0.110$ \\
Risk Manager          & $1.000 \pm 0.000$ & $0.962 \pm 0.196$ & $0.962 \pm 0.196$ \\
CSCO                  & $0.950 \pm 0.201$ & $0.893 \pm 0.288$ & $0.899 \pm 0.277$ \\
\bottomrule
\end{tabular}%
}
\end{table}

The evaluation results demonstrate strong performance across all agents, with several agents achieving near-perfect or perfect precision scores. These high scores reflect both the framework's architectural design and deliberate prompt engineering strategies. The framework employs strict JSON output specifications in agent prompts, requiring agents to produce outputs in predefined schemas. This approach facilitates deterministic tool integration and enables rigorous evaluation. The prompt engineering approach, combined with deterministic tool grounding for critical computations, creates a hybrid architecture where LLM-based reasoning orchestrates reliable computational components.

The agentic behaviour in this framework manifests through autonomous tool selection, parameter extraction from unstructured inputs, error detection and recovery, and context-aware decision-making. The framework deliberately grounds critical computations in deterministic tools to ensure reliability and reproducibility. This design choice is essential for supply chain risk assessment, where computational errors can lead to significant operational and financial consequences. This design pattern reflects best practices in production AI systems, where deterministic grounding provides auditability and regulatory compliance while LLM orchestration enables flexibility and adaptability to diverse disruption scenarios.

\subsubsection{Disruption Monitoring Agent: Foundation for Agentic Pipeline}

The Disruption Monitoring Agent achieves an F1 score of $0.991 \pm 0.028$ with precision of $0.983 \pm 0.051$ and recall of $1.000 \pm 0.000$. This performance is critical because the agent serves as the foundation of the entire framework, transforming unstructured news text into structured disruption entities that downstream agents require for knowledge graph traversal and risk assessment. The perfect recall ($1.000 \pm 0.000$) indicates that the agent successfully identifies all entities mentioned in ground truth, ensuring that no critical disruption information is lost at the initial extraction stage. The high precision ($0.983 \pm 0.051$) indicates that false positives are rare, which is essential for reducing false alarms in production deployments where incorrect entity extraction would trigger unnecessary supply chain investigations.

The agent's high performance reflects several design choices. First, strict JSON output specifications in prompts enforce structured entity extraction. Second, explicit instructions require extraction of only entities directly mentioned in the text. The performance is inherently limited by the information content of input text. Explicit entity mentions yield perfect scores, while implicit references show lower performance. This behaviour is expected and appropriate for information extraction systems, where the agent can only extract information that exists in the source material. The near-perfect performance across diverse disruption types demonstrates the agent's robustness and its capability to serve as a reliable foundation for the proposed agentic framework.

This performance level means that the framework can reliably identify disruption events affecting specific countries, industries, and companies from news sources, enabling automated monitoring of global supply chain disruptions. The perfect recall ensures that critical disruptions are not missed. The high precision minimises false alarms that would waste organisational resources. The agent's role as the entry point makes its accuracy particularly important, as errors propagate through subsequent agents, potentially leading to incorrect risk assessments and strategic decisions.

\subsubsection{Knowledge Graph Query Agent: Multi-Tier Network Traversal}

The Knowledge Graph Query Agent achieves an average F1 score of $0.980 \pm 0.110$ with precision of $1.000 \pm 0.000$ and recall of $0.975 \pm 0.137$. The perfect precision ($1.000 \pm 0.000$) indicates zero false positives, meaning that every disrupted path identified by the agent is correct. This precision is achieved through the agent's orchestration of deterministic Neo4j Cypher queries. The agent extracts disruption entities from upstream outputs, formats graph traversal queries, and executes them against the knowledge graph. The deterministic nature of Neo4j queries ensures computational reliability: given the same disruption entities, the same paths are always identified, eliminating variability in path discovery. The agent's role is to correctly extract parameters, format queries, and interpret results, which it accomplishes with perfect precision when given correct inputs.

The high recall ($0.975 \pm 0.137$) indicates that the agent successfully identifies the vast majority of disrupted paths across multi-tier supply chains. This capability is fundamental to the framework's objective of comprehensive risk assessment, as missing disrupted paths would lead to underestimated risk scores for Tier-1 suppliers. The standard deviation of 0.137 reflects that most scenarios achieve perfect or near-perfect recall when given correct inputs, while occasional misses occur due to cascading errors from upstream entity extraction failures. This performance means that the framework can reliably identify all supply chain paths affected by a disruption, enabling a comprehensive risk assessment that accounts for both direct and indirect supplier dependencies.

\subsubsection{Risk Manager Agent: Tier-1 Risk Quantification}

The Risk Manager Agent achieves an averaged precision of $1.000 \pm 0.000$, recall of $0.962 \pm 0.196$, and F1 score of $0.962 \pm 0.196$. The perfect precision ($1.000 \pm 0.000$) is achieved through the agent's orchestration of deterministic risk calculation tools provided to the agent. These tools aggregate downstream disruption impacts into Tier-1 supplier risk scores. The agent uses the tool provided to perform mathematical calculations to compute four risk dimensions: exposure depth (maximum disrupted tier), exposure breadth (count of disrupted downstream suppliers), downstream criticality (aggregated criticality of disrupted suppliers), and network centrality (graph-based centrality measures). These calculations are deterministic, ensuring that given the same disrupted paths, the same risk scores are always calculated. The agent's role is to orchestrate these calculations, filter to the top 10 riskiest suppliers, and structure the output according to strict JSON schemas specified in prompts.

The recall of $0.962 \pm 0.196$ indicates that the agent successfully identifies 96\% of the top 10 riskiest suppliers on average. This performance is directly dependent on the Knowledge Graph Query Agent's completeness. When upstream agents correctly identify disrupted paths, the Risk Manager Agent achieves perfect or near-perfect recall. When upstream agents fail (due to incorrect entity extraction), the Risk Manager Agent has no paths to calculate risk scores for, resulting in zero suppliers identified. This cascading dependency reflects the agentic architecture's behaviour, where each agent's output becomes the next agent's input.

\subsubsection{CSCO Agent: Strategic Decision-Making}
\label{subsec:csco_quantitative}

The CSCO Agent achieves a precision of $0.950 \pm 0.201$, recall of $0.893 \pm 0.288$, and F1 score of $0.899 \pm 0.277$. {This agent transforms quantitative risk assessments into strategic decisions, addressing the framework's ultimate objective of providing actionable supply chain intelligence.} {The agent's decision-making process involves both deterministic rule application (mapping risk scores to actions using predefined thresholds) and agentic reasoning (interpreting risk profiles, considering company context, generating executive-level justifications).} {The higher precision (0.950) compared to recall (0.893) indicates that when the agent makes decisions, they are highly accurate, but the agent occasionally misses some suppliers that should be addressed.} {This reflects the complexity of strategic decision-making where multiple factors beyond risk scores may influence decisions.}

In the section below, we conduct a further qualitative evaluation of the CSCO agent to assess completeness, clarity, relevance and insightfulness of the agent outputs.

\subsubsection{Qualitative Evaluation of CSCO Agent}
\label{subsec:csco_qualitative}

In addition to quantitative metrics, we evaluate the CSCO Agent's qualitative analysis sections using a rubric-based methodology. {The qualitative evaluation assesses three sections: disruption summary, network impact analysis, and replacement recommendations.} {Each section is evaluated by human evaluators. following weighted criteria, as specified in \autoref{tab:qualitative_rubric}.}

\begin{table}[h]
\centering
\caption{Qualitative Evaluation Rubric for CSCO Agent}
\label{tab:qualitative_rubric}
\resizebox{0.6\textwidth}{!}{%
\begin{tabular}{lcc}
\toprule
\textbf{Section} & \textbf{Criterion} & \textbf{Weight} \\
\midrule
\multirow{3}{*}{Disruption Summary} & Completeness & 0.4 \\
 & Clarity & 0.3 \\
 & Relevance & 0.3 \\
\midrule
\multirow{3}{*}{Network Impact Analysis} & Completeness & 0.4 \\
 & Accuracy & 0.4 \\
 & Insightfulness & 0.2 \\
\midrule
\multirow{3}{*}{Replacement Recommendations} & Completeness & 0.4 \\
 & Accuracy & 0.4 \\
 & Actionability & 0.2 \\
\bottomrule
\end{tabular}%
}
\end{table}

{The disruption summary is evaluated for completeness (covers what happened, who is affected, why it is important to Tier-0, and operational impact), clarity (professional, executive-level language), and relevance (accurately reflects disruption type and affected entities).} {The network impact analysis is evaluated for completeness (mentions companies affected, tier distribution, propagation likelihood), accuracy (accurately reflects knowledge graph results), and insightfulness (provides meaningful analysis beyond raw numbers).} {The replacement recommendations are evaluated for completeness (identifies suppliers to replace, provides justification, explains impact), accuracy (recommendations align with actual decisions and risk scores), and actionability (provides clear, actionable guidance).}

{Table~\ref{tab:qualitative_results} presents the qualitative evaluation results aggregated across all scenarios.} {The CSCO Agent achieves strong performance across all three sections, with weighted mean scores of $0.811 \pm 0.049$ for disruption summary, $0.486 \pm 0.172$ for network impact analysis, and $0.830 \pm 0.078$ for replacement recommendations, resulting in an overall qualitative score of $0.709 \pm 0.099$.} {These scores indicate that the agent provides comprehensive, accurate, and actionable qualitative analysis that supports executive-level decision-making.}

\begin{table}[h]
\centering
\caption{Qualitative Evaluation Results for CSCO Agent (Mean $\pm$ Std)}
\label{tab:qualitative_results}
\resizebox{\textwidth}{!}{%
\begin{tabular}{lcccc}
\toprule
\textbf{Section} & \textbf{Completeness} & \textbf{Clarity/Accuracy} & \textbf{Relevance/Insight/Action} & \textbf{Weighted Mean} \\
\midrule
Disruption Summary & $0.777 \pm 0.077$ & $0.893 \pm 0.058$ & $0.773 \pm 0.097$ & $0.811 \pm 0.049$ \\
Network Impact Analysis & $0.637 \pm 0.155$ & $0.317 \pm 0.251$ & $0.523 \pm 0.139$ & $0.486 \pm 0.172$ \\
Replacement Recommendations & $0.857 \pm 0.103$ & $0.763 \pm 0.092$ & $0.910 \pm 0.111$ & $0.830 \pm 0.078$ \\
\midrule
\textbf{Overall Qualitative Score} & \multicolumn{4}{c}{\textbf{$0.709 \pm 0.099$}} \\
\bottomrule
\end{tabular}%
}
\end{table}

{The qualitative evaluation demonstrates that the CSCO Agent provides executive-level communication quality with varying performance across sections.} {The disruption summary achieves strong performance ($0.811 \pm 0.049$), indicating that the agent effectively communicates what happened, who is affected, and why it matters to the monitored company.} {The high clarity score ($0.893 \pm 0.058$) reflects professional, executive-level language.} {The network impact analysis shows lower performance ($0.486 \pm 0.172$), with accuracy being the primary challenge ($0.317 \pm 0.251$).} {This reflects difficulties in accurately reflecting knowledge graph results in the qualitative text, potentially due to the complexity of translating long extensive quantitative path data into narrative form.} {The replacement recommendations achieve the strongest performance ($0.830 \pm 0.078$), with particularly high actionability ($0.910 \pm 0.111$), indicating that the agent provides clear, actionable guidance aligned with risk assessments.} {This qualitative performance complements the quantitative metrics, providing a comprehensive assessment of the agent's reasoning capabilities and communication quality.} {The lower network impact analysis scores suggest an area for improvement in translating quantitative knowledge graph results into accurate qualitative narratives.} {The high actionability scores indicate that the agent provides decision-makers with clear, actionable guidance, which is essential for operational response to disruptions.}

\subsubsection{Error Propagation and Framework Robustness}

{Analysis of performance across the agentic pipeline reveals important insights into error propagation and framework robustness.} {The framework exhibits a cascading error pattern where Disruption Monitoring Agent errors propagate through subsequent agents.} {When entity extraction fails, the KG Query Agent uses incorrect parameters, resulting in zero disrupted paths identified.} {This causes the Risk Manager Agent to have no paths to calculate risk scores for, resulting in zero suppliers identified.} {This prevents the CSCO Agent from making decisions.} {This cascading effect highlights the critical importance of accurate information extraction as the foundation for the entire framework.} {Future improvements should focus on enhancing entity extraction robustness, potentially through ensemble methods or confidence-based filtering.}

{However, when the Disruption Monitoring Agent succeeds, downstream agents generally achieve high performance.} {The KG Query Agent achieves perfect precision.} {The Risk Manager Agent achieves perfect precision and high recall.} {The CSCO Agent achieves high F1 scores.} {This demonstrates that the framework's architecture successfully isolates errors to the initial extraction stage, preventing error amplification through deterministic tool grounding.} {The deterministic tools ensure that computational errors do not compound.} {The agentic orchestration provides flexibility for handling diverse scenarios.} {This design pattern is particularly important for supply chain applications, where computational reliability is essential for operational decision-making, yet flexibility is needed to handle the diverse and unpredictable nature of supply chain disruptions.}

\subsubsection{System Runtime and Computational Requirements}

\rev{To assess the practical feasibility of the proposed framework, we evaluate its runtime performance and computational costs across all 30 evaluation scenarios.} \rev{This analysis provides insights into the framework's operational efficiency and cost-effectiveness for real-world deployment.}

\rev{Table~\ref{tab:runtime_metrics} presents execution time and cost statistics aggregated across all 30 scenarios.} \rev{The system was executed on a 16-inch MacBook Pro (Apple M3 Max, 48GB RAM).} \rev{The full end-to-end workflow, from disruption detection and risk propagation through to supplier identification and mitigation recommendation, completed in a mean time of 3.83 minutes per scenario.} \rev{This includes agent-to-agent communication, LLM reasoning, knowledge graph queries, risk calculations, and external data retrieval.} \rev{Execution times range from 1.67 minutes for simple scenarios with minimal affected paths to 6.78 minutes for complex scenarios with extensive multi-tier disruptions, with a standard deviation of 1.23 minutes.}

\begin{table}[h]
\centering
\caption{System Runtime and Cost Metrics Across 30 Evaluation Scenarios}
\label{tab:runtime_metrics}
\resizebox{0.7\textwidth}{!}{%
\begin{tabular}{lcc}
\toprule
\textbf{Metric} & \textbf{Execution Time} & \textbf{Cost (USD)} \\
\midrule
Mean & 3.83 min  & \$0.0836 \\
Median & 4.01 min  & \$0.0765 \\
Standard Deviation & 1.23 min & \$0.0263 \\
Minimum & 1.67 min & \$0.0348 \\
Maximum & 6.78 min & \$0.1251 \\
\midrule
\textbf{Total (30 scenarios)} & \textbf{115.0 min } & \textbf{\$2.51} \\
\bottomrule
\end{tabular}%
}
\end{table}

\rev{In comparison, most human-led supply chain teams typically take five days on average to respond to disruptions, according to a global survey of 1,800 supply chain leaders \citep{kinaxis2024idc}.} \rev{This demonstrates that our autonomous framework reduces response time by nearly three orders of magnitude, dramatically improving situational awareness and enabling proactive intervention.} \rev{This time reduction is especially important when disruptions originate beyond Tier-1 suppliers, where manual investigation of deep-tier dependencies can take weeks.} \rev{The framework's ability to complete comprehensive multi-tier analysis in minutes rather than days enable enables organisations to respond before disruptions cascade and impact production operations.}

\rev{The framework utilizes OpenAI's GPT-4o model for all natural language processing and reasoning tasks.} \rev{According to current pricing, GPT-4o costs \$5.00 per million input tokens and \$15.00 per million output tokens \citep{openai2024pricing}.} \rev{The mean cost per scenario across all 30 evaluations was \$0.0836 USD.} \rev{This demonstrates the framework's viability for repeated, real-world use, as the cost per disruption analysis is minimal compared to the value of early detection and mitigation.} \rev{Costs scale approximately with token volume and model choice.} \rev{Longer source articles, richer prompts, or higher verbosity settings increase usage; conversely, prompt compaction and output-length constraints reduce it.} \rev{Concurrency (multiple simultaneous cases) lifts throughput but can increase spend unless caching and batching are applied.}

\paragraph*{Web Search Integration} 
\rev{To identify potential alternative suppliers and verify product availability, the system integrates with SerpAPI (a web search tool) \citep{serper2025pricing}.} \rev{While SerpAPI offers a free usage tier, additional charges apply based on query volume and concurrency.} \rev{In our deployment across 30 scenarios, usage remained within the free tier, incurring no additional cost.}

\paragraph*{Scalability} 
\rev{The modular design supports parallel execution, but we have not yet stress-tested the system under high load.} \rev{Key bottlenecks are (i) LLM concurrency limits and token throughput, (ii) external API rate limits, and (iii) graph-query fan-out on large, deep supplier networks.} \rev{Practical mitigations include prompt/output truncation, response caching across agents, batched graph traversals, and asynchronous job queues.} \rev{A formal scale test (throughput, tail latency, cost/alert at increasing case volumes) is left to future work.}

\subsection{Limitations and Future Research Directions}

\rev{While the framework demonstrates high performance across 30 evaluation scenarios, several limitations should be acknowledged: (1) the framework currently requires manual article ingestion rather than integration with real-time news APIs, limiting its operational mode to batch processing rather than continuous monitoring; (2) evaluation is conducted on three automotive manufacturers, and broader validation across different industries (electronics, pharmaceuticals, aerospace) with larger datasets would strengthen generalisability; (3) the knowledge graph represents a static snapshot of supply chain relationships, whereas real networks evolve continuously, and temporal network data would enhance long-term accuracy; (4) the system has not been stress-tested under high load or concurrent execution scenarios, leaving scalability characteristics unvalidated for production-scale deployment.}

\rev{Future research will focus on six critical areas: (1) integration with live news APIs and real-time data streams, requiring research on streaming data processing and temporal reasoning; (2) validation on larger datasets of real-world disruption events across multiple industries to strengthen generalisability; (3) development of temporal network representations that capture supply chain evolution over time to enhance framework accuracy for long-term deployment; (4) systematic scalability testing under high load, concurrent execution, and increasing case volumes to establish performance boundaries and inform production deployment strategies; (5) comparative evaluation against existing commercial solutions and ablation studies isolating the contribution of agentic orchestration versus deterministic tool execution to quantify the added value of the proposed architecture; (6) formal human-in-the-loop evaluation studies assessing how supply chain managers interact with and validate agent recommendations in operational settings, measuring decision quality, trust calibration, and time-to-action improvements.  } 

In the following \autoref{section:case-study}, we present a full walkthrough of each agent's outputs using a real-world disruption scenario. We use the Russia--Ukraine war as a representative geopolitical disruption. This case study illustrates how the framework operates in practice and the type of insights each agent produces across the disruption monitoring pipeline.}

\section{Case Study: Russia--Ukraine War} \label{section:case-study}

\rev{This section presents a walkthrough of a real-world disruption scenario to illustrate the framework's operational value and managerial implications.} Specifically, we apply the system as if it were integrated within the supply chain risk management operations of \textit{Mercedes-Benz Group AG}, using the 2022 escalation of the Russia--Ukraine War as the test disruption event. This scenario was selected for its global economic relevance, strong supply chain implications, and well-documented impact on critical materials.

\begin{figure}[H]
\centering
    \includegraphics[width=1\linewidth]{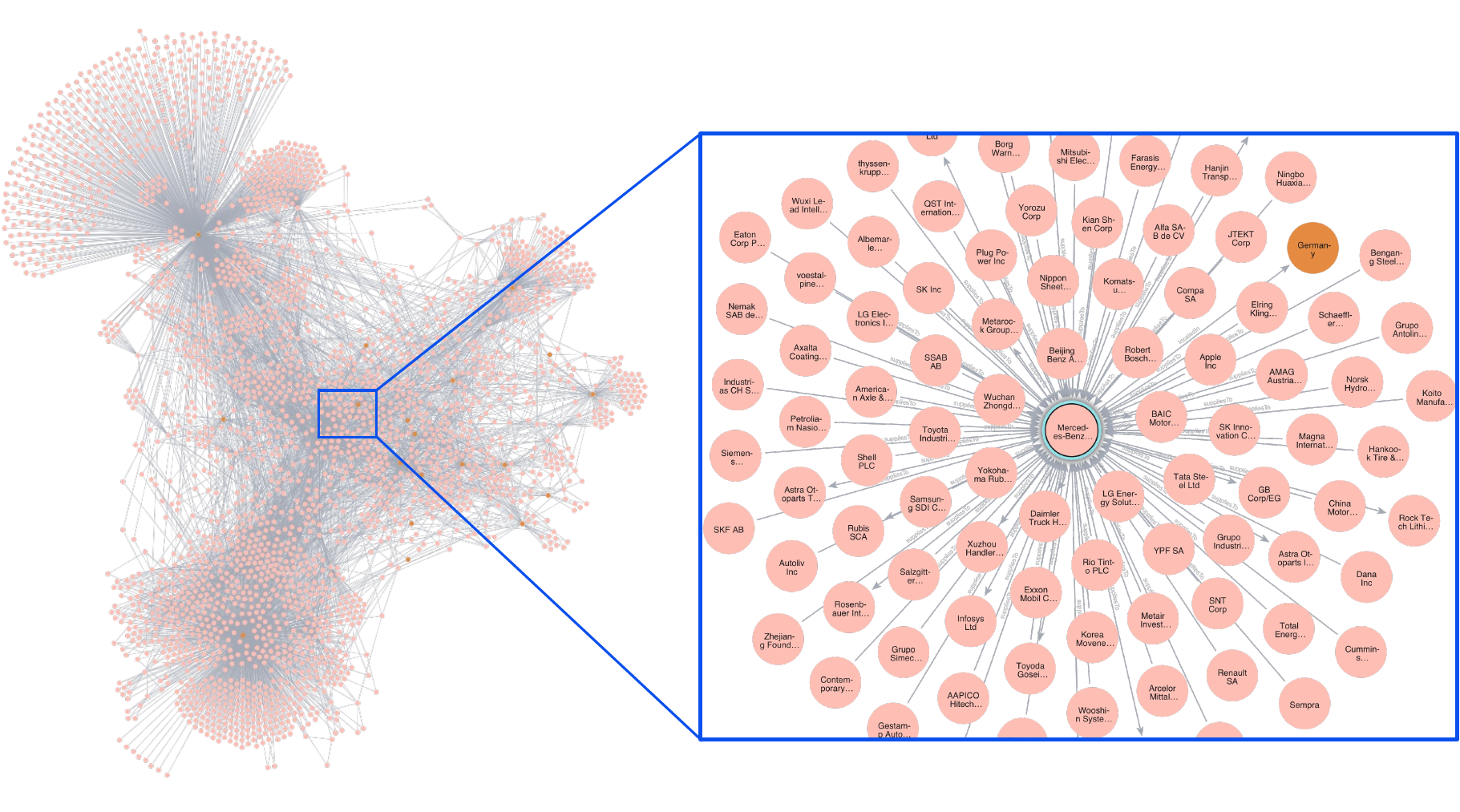}
    \caption{A snapshot of the knowledge graph showing Mercedes-Benz Group AG's multi-tier supplier and country relationships, used by the Knowledge Graph Query Agent to detect and trace disruption impacts across its extended supply chain.}
    
\label{fig:knowledge-graph}
\end{figure}

In this case study, we simulate a realistic operational pipeline in which a single news article is manually selected for its relevance to the disruption, is fed into the system. This choice ensures consistency across agent prompts and allows us to validate outputs in a structured manner. The Disruption Monitoring Agent is prompted as if it were a senior risk analyst at Mercedes-Benz, and all downstream agents operate under corresponding role-based assumptions. For this experiment, we have access to a pre-constructed, company-specific knowledge graph of Mercedes-Benz's extended supply chain. This graph includes verified Tier-1 to Tier-4 relationships and was built in a previous study \citep{almahri2024enhancing}; a relevant subgraph used in this scenario is illustrated in \autoref{fig:knowledge-graph}.

 To evaluate the accuracy of each agent, all outputs were manually reviewed to ensure alignment with the expected behaviour defined in agent prompts. The following sections present a step-by-step walkthrough of each agent's output and behaviour in the simulated scenario.

\subsection{Disruption Monitoring Agent}

\begin{figure}[H]
  \centering
  \includegraphics[width=\linewidth]{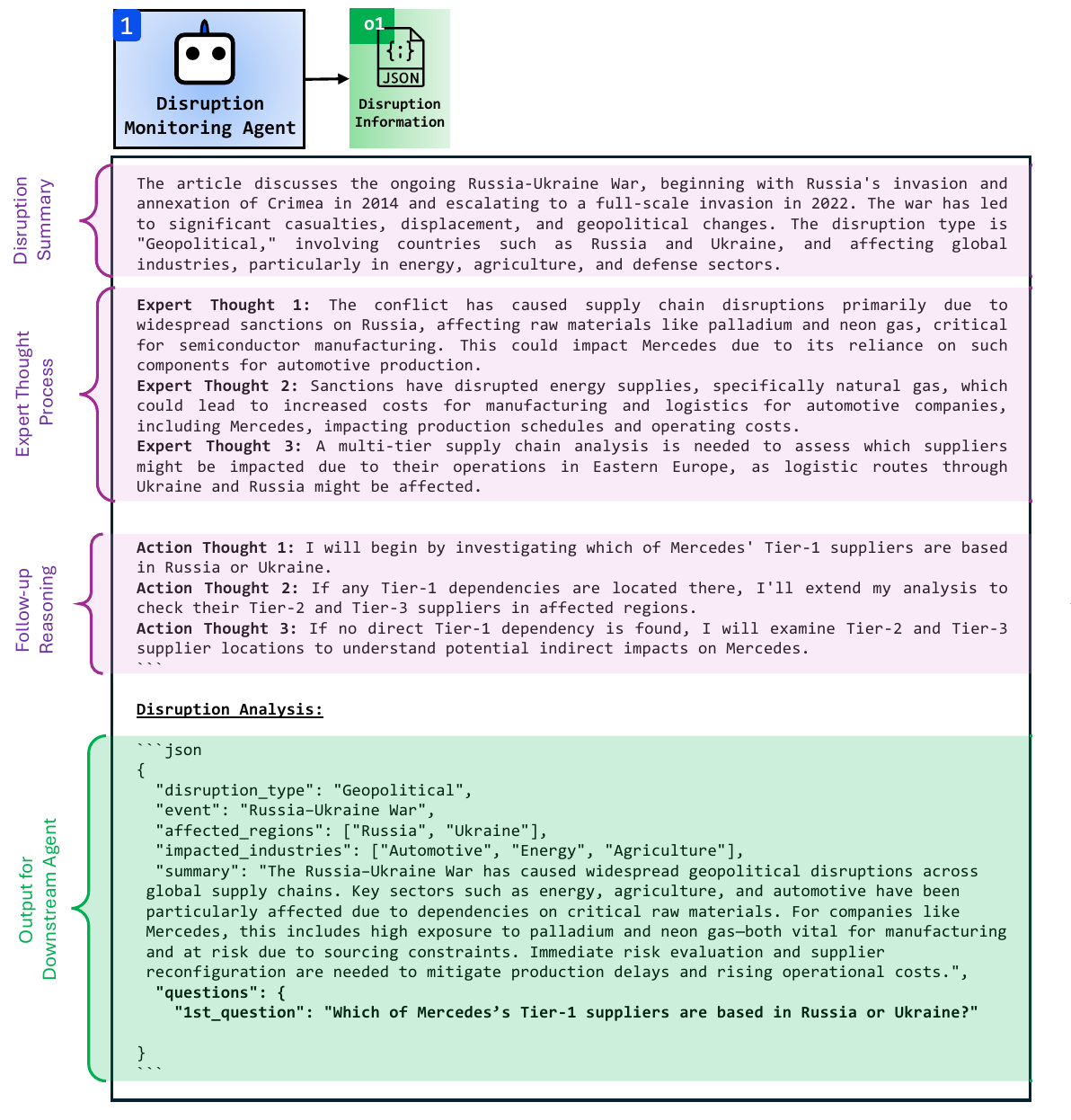}
  \caption{Full JSON output of the Disruption Monitoring Agent. The agent correctly classifies the geopolitical event, extracts key entities and industries, reasons through expert-level insights, and generates precise diagnostic questions for the knowledge graph query stage.}
  \label{fig:scenario1-agent1}
\end{figure}

\rev{The Disruption Monitoring Agent processes a news article covering the 2022 Russia--Ukraine War and generates structured disruption intelligence, as shown in \autoref{fig:scenario1-agent1}.} \rev{The agent correctly classifies the event as a geopolitical disruption and identifies affected countries (Russia, Ukraine) and industries (Metals \& Mining, Energy, Chemicals).} \rev{It produces three expert-level insights: (1) potential disruptions to palladium and neon gas critical for semiconductor manufacturing, (2) indirect impacts of energy shortages from sanctions affecting cost and delivery timelines, and (3) geographical risks requiring deeper investigation of Eastern European suppliers.}

\subsection{Knowledge Graph Query Agent}

\begin{figure}[H]
  \centering
  \includegraphics[width=\linewidth]{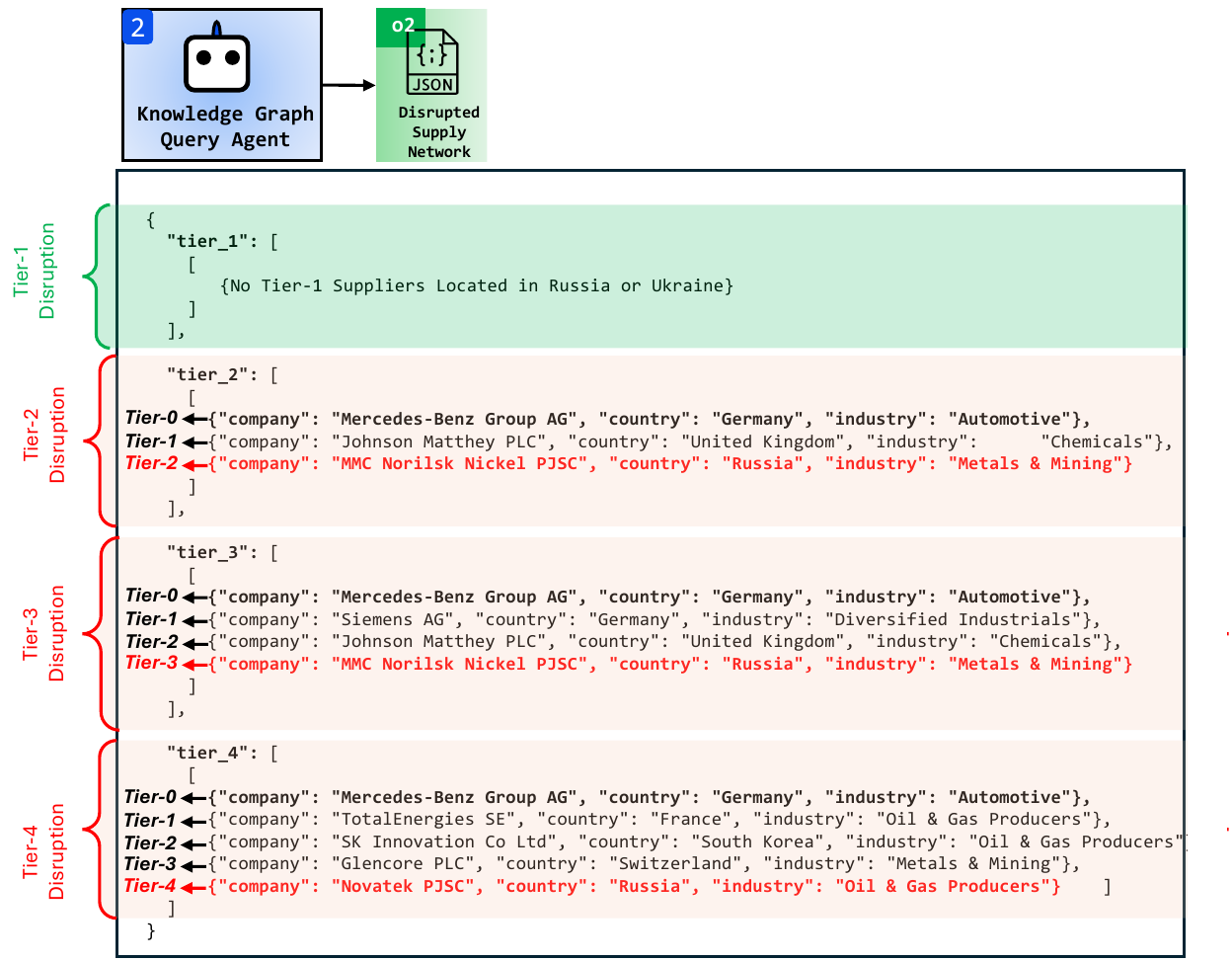}
  \caption{Partial JSON output of the Knowledge Graph Query Agent. The agent resolves Mercedes-Benz's suppliers up to Tier-4 in Russia and Ukraine, returning complete, tier-annotated dependency chains in strict JSON format.}
  \label{fig:scenario1-agent2}
\end{figure}

\rev{The Knowledge Graph Query Agent receives the structured disruption entities and identifies all affected suppliers across Mercedes-Benz's extended supply chain, as shown in \autoref{fig:scenario1-agent2}.} \rev{The agent reveals that while no Tier-1 suppliers are directly located in Russia or Ukraine, critical dependencies exist at deeper tiers.} \rev{Specifically, Johnson Matthey PLC, a Tier-1 supplier in the chemicals sector, depends on MMC Norilsk Nickel PJSC, a major Russian mining firm, creating a Tier-2 disruption pathway.} \rev{The agent also identifies Tier-3 and Tier-4 dependencies involving Siemens AG and TotalEnergies SE, which are indirectly linked to Russian suppliers including Novatek PJSC and Glencore PLC.}

\rev{The framework completes this analysis in minutes, automatically traversing the knowledge graph to identify all affected paths up to Tier-4.} \rev{For Mercedes-Benz, this immediate visibility into the Johnson Matthey--Norilsk Nickel dependency enables proactive mitigation before the disruption cascades, potentially preventing production delays and material shortages that could cost millions in lost revenue.} 

\subsection{Product Search Agent}

\begin{figure}[H]
  \centering
  \includegraphics[width=\linewidth]{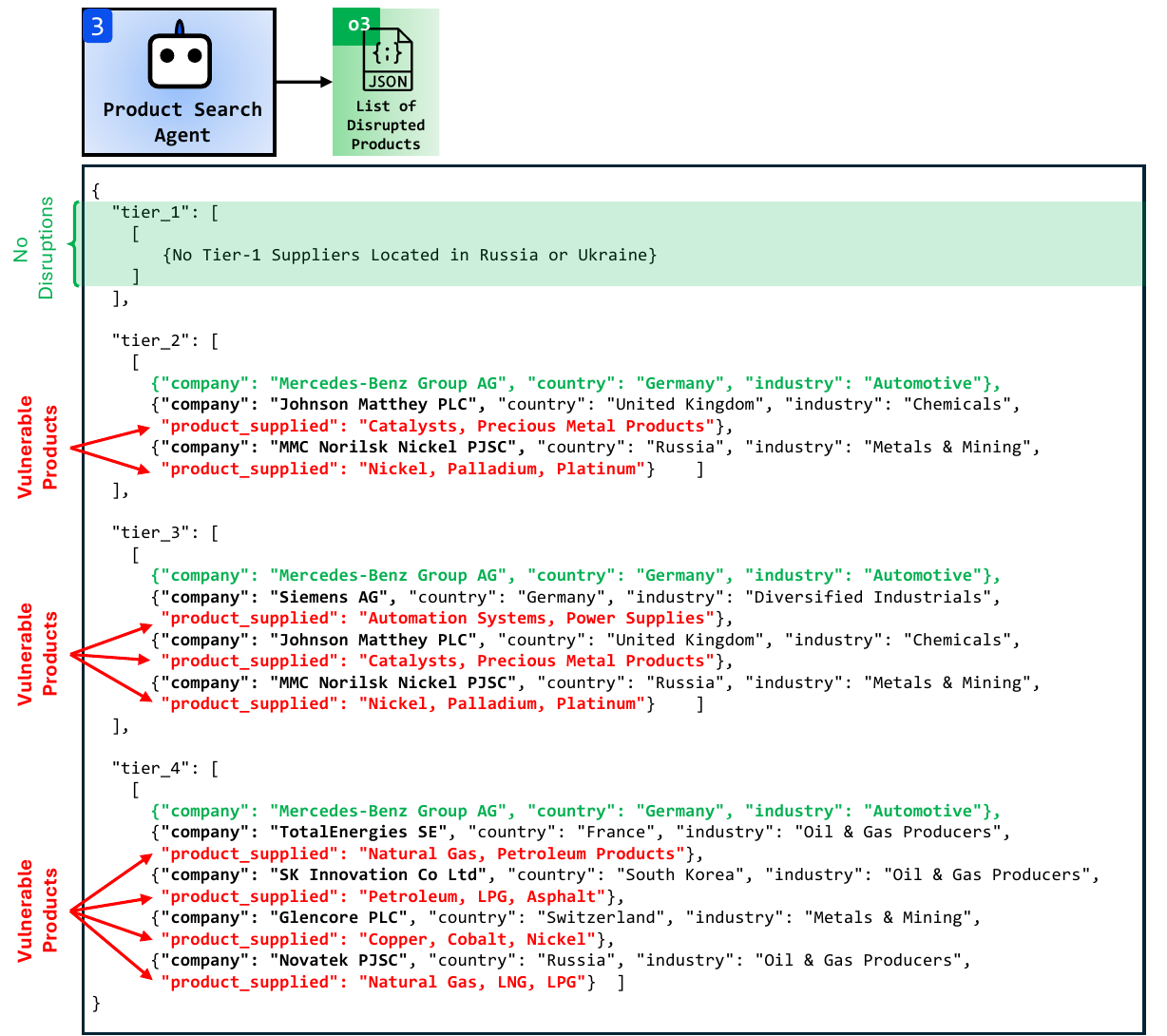}
  \caption{Full JSON output of the Product Traceability Agent. Each disrupted supplier chain is augmented with the specific products supplied, facilitating material-specific impact analysis (e.g.\ catalysts and palladium for automotive applications).}
  \label{fig:scenario1-agent3}
\end{figure}

\begin{figure}[H]
  \centering
  \includegraphics[width=\linewidth]{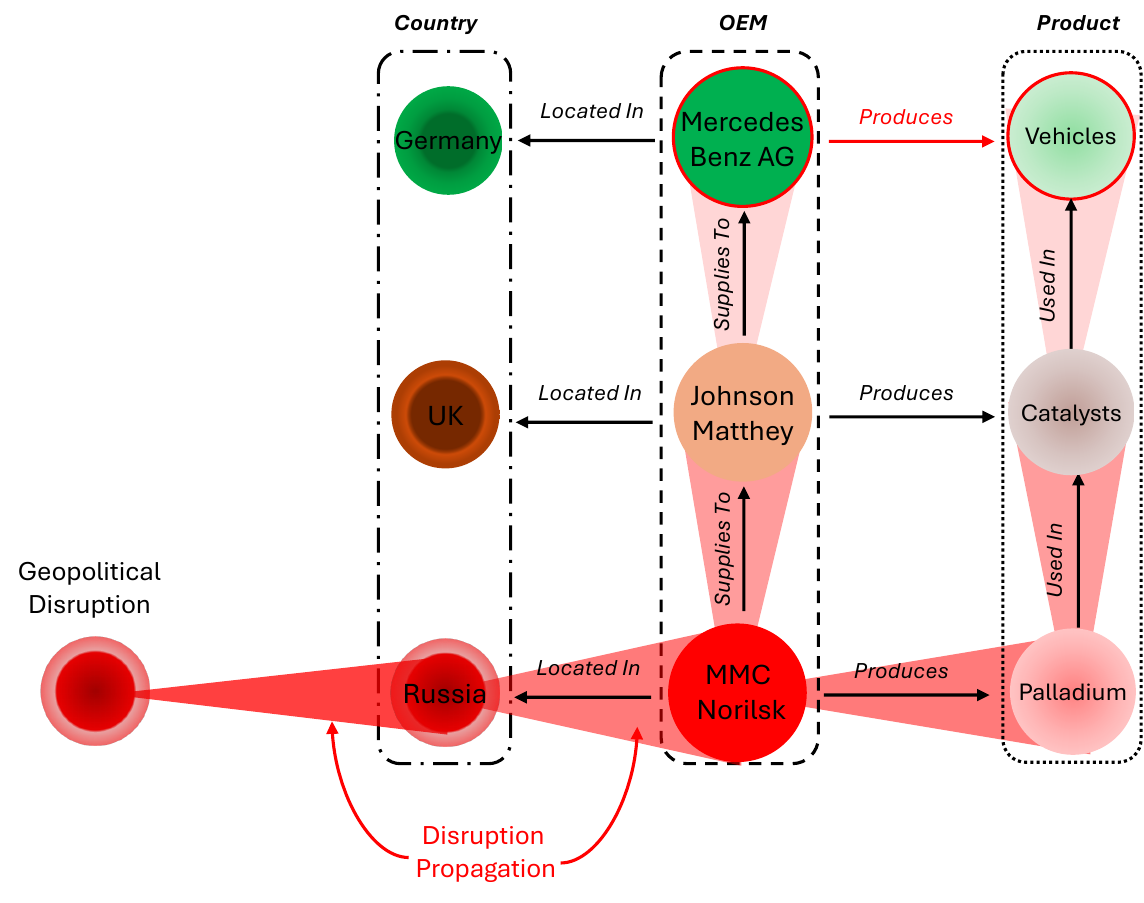}
  \caption{Annotated supply chain subnetwork produced by the Product Search Agent. The figure illustrates the disruption propagation from MMC Norilsk Nickel PJSC (Russia) through Johnson Matthey PLC (UK) to Mercedes-Benz Group AG (Germany). Each node is enriched with product-level information, revealing that palladium produced in Russia is used in catalysts supplied to Mercedes-Benz, exposing a critical material dependency relevant to catalytic converter manufacturing.}

  \label{fig:scenario1-agent3-2}
\end{figure}

\rev{The Product Search Agent enriches the disrupted supplier chains with product-level information, as shown in \autoref{fig:scenario1-agent3} and \autoref{fig:scenario1-agent3-2}.} \rev{The agent associates Johnson Matthey PLC with "Catalysts, Precious Metal Products" and MMC Norilsk Nickel PJSC with "Nickel, Palladium, Platinum," creating a complete material flow traceability from raw materials to finished components.}

\rev{This product-level traceability transforms generic supplier risk into actionable material-specific intelligence.} \rev{For Mercedes-Benz, the framework reveals that catalytic converter production depends on palladium sourced from Russia through the Johnson Matthey--Norilsk Nickel pathway.} \rev{This specificity enables targeted mitigation strategies: procurement teams can focus on securing alternative palladium sources rather than investigating all supplier relationships.} \rev{In practice, this reduces the scope of mitigation efforts from potentially hundreds of supplier relationships to specific material flows, dramatically reducing the time and resources required for response planning.} \rev{The ability to trace materials from raw sources through intermediate processing to final components enables more precise inventory planning, alternative sourcing decisions, and production scheduling adjustments that minimise operational disruption.}

\subsection{Network Visualization Agent}

\begin{figure}[H]
  \centering
  \includegraphics[width=\linewidth]{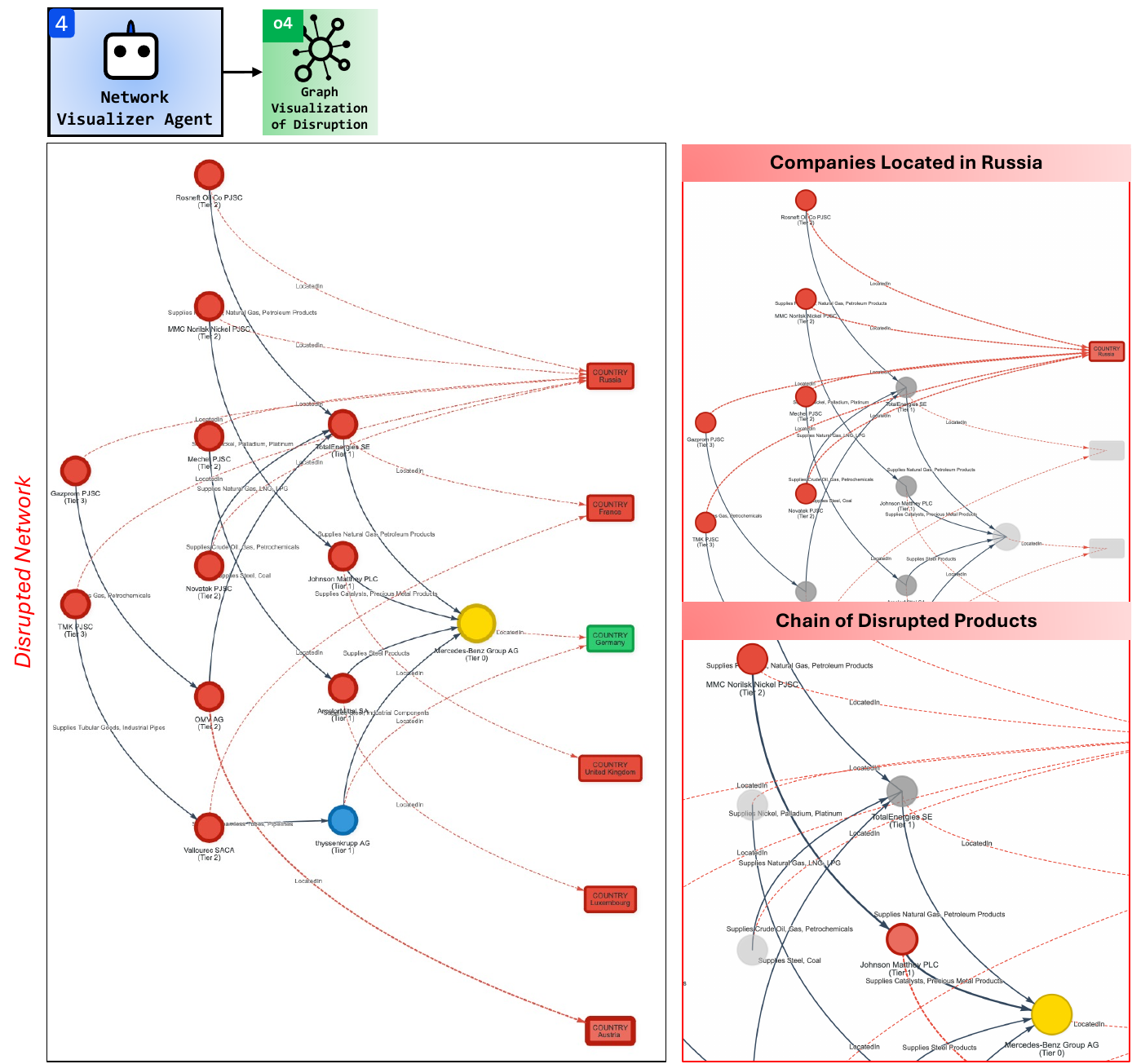}
  \caption{Network Visualization Agent output. The disrupted supply network is plotted with colour-coded risk levels (red for high exposure), and edge labels indicating key material flows, all generated automatically from the agent's prompt.}
  \label{fig:scenario1-agent4}
\end{figure}

\rev{The Network Visualization Agent generates an interactive network diagram (\autoref{fig:scenario1-agent4}) that visually represents the disrupted supply network, highlighting critical nodes such as MMC Norilsk Nickel PJSC and Johnson Matthey PLC in red.} \rev{The visualization traces disruption pathways from affected suppliers back to Mercedes-Benz, making risk propagation immediately apparent.}

\rev{Visual representations accelerate decision-making by enabling executives and supply chain managers to quickly understand disruption scope and prioritise response actions.} \rev{Traditional risk reports require extensive reading and analysis to extract key insights, often taking hours for decision-makers to fully comprehend the situation.} \rev{The framework's automated visualization presents the same information in a format that can be understood in minutes, enabling faster executive briefings and more rapid approval of mitigation strategies.} \rev{This time compression is critical during disruption events, where delays in decision-making can result in material shortages, production stoppages, and revenue losses.} \rev{The visual format also facilitates communication across organizational functions, enabling procurement, operations, and executive teams to align on response priorities without extensive briefing sessions.}

\subsection{Risk Manager Agent}

\begin{figure}[H]
  \centering
  \includegraphics[width=0.5\linewidth]{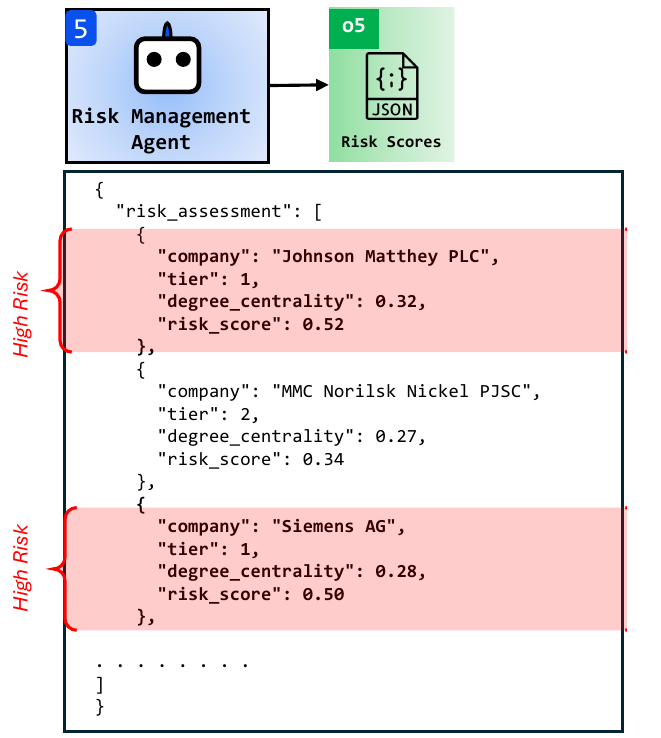}
  \caption{Partial JSON output of the Risk Assessment Agent. Each supplier is listed with its tier, normalized degree centrality, and composite risk score, computed as a weighted sum of centrality (70\%) and inverse-tier weight (30\%).}
  \label{fig:scenario1-agent5}
\end{figure}

\rev{The Risk Manager Agent calculates comprehensive risk scores for Tier-1 suppliers by aggregating exposure depth, exposure breadth, downstream criticality, and network centrality metrics, as shown in \autoref{fig:scenario1-agent5}.} \rev{The agent identifies Johnson Matthey PLC and Siemens AG as the highest-risk suppliers due to their Tier-1 position, strong network connectivity, and exposure to disrupted downstream suppliers.}

 \rev{For Mercedes-Benz, identifying Johnson Matthey as the highest-risk supplier enables immediate allocation of procurement resources to secure alternative palladium sources, rather than investigating all suppliers equally.}

\subsubsection{CSCO Agent}

\begin{figure}[H]
  \centering
  \includegraphics[width=\linewidth]{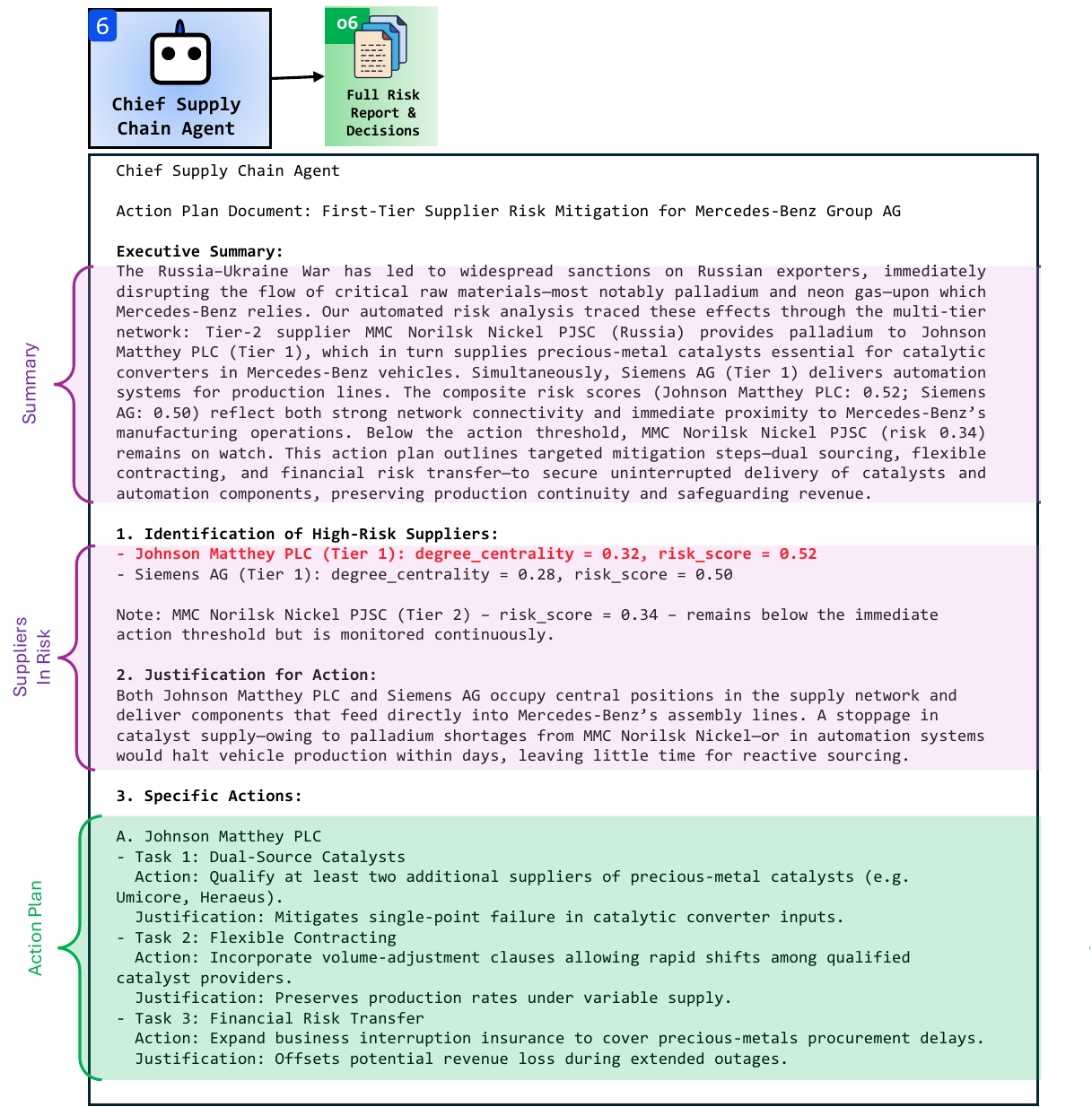}
  \caption{Partial action plan output of the Chief Supply Chain Agent for the Russia--Ukraine War scenario}
  \label{fig:scenario1-agent6}
\end{figure}

\rev{The CSCO Agent synthesizes all upstream analytics into an executive-level action plan (\autoref{fig:scenario1-agent6}), recommending specific mitigation strategies for high-risk suppliers.} \rev{The agent identifies Johnson Matthey PLC and Siemens AG as priority suppliers requiring immediate attention and recommends actions such as dual-sourcing of catalysts, flexible contracting, and expanded financial risk transfer mechanisms.}

\rev{This automated synthesis addresses a critical bottleneck in disruption response: translating analytical findings into actionable executive decisions.} \rev{Traditional processes require risk analysts to compile findings, draft recommendations, and prepare executive briefings, typically taking 1-2 days.} \rev{The framework generates a comprehensive action plan in minutes, including quantitative risk justification, recommended actions, and implementation priorities.} \rev{For supply chain managers, this immediate availability of structured recommendations enables faster decision-making and reduces the time from disruption detection to mitigation action.} \rev{The executive-level format also facilitates communication with C-suite executives, enabling rapid approval of resource allocation and strategic decisions that might otherwise require multiple briefing cycles.} \rev{The human-in-the-loop review process ensures that recommendations align with corporate strategy and risk tolerance, while the automated generation eliminates the time-consuming manual preparation of decision documents.}

\subsubsection{Alternative Sourcing Agent}

\begin{figure}[H]
  \centering
  \includegraphics[width=\linewidth]{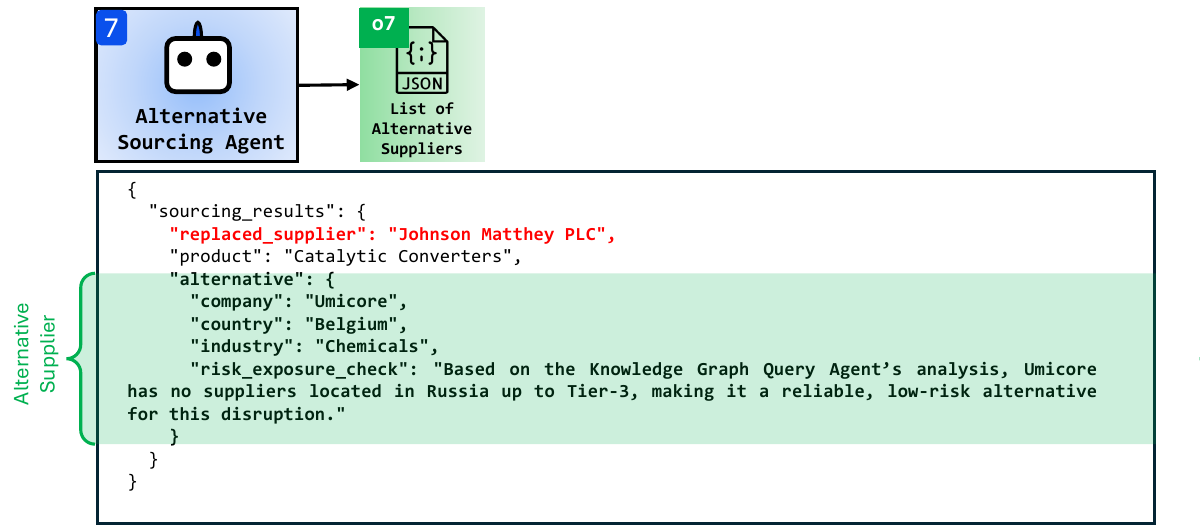}
  \caption{Alternative sourcing agent output}
  \label{fig:scenario1-agent7}
\end{figure}

\rev{The Alternative Sourcing Agent identifies Umicore, a Belgium-based chemicals firm, as a viable alternative to Johnson Matthey PLC, as shown in \autoref{fig:scenario1-agent7}.} \rev{The agent validates this selection by querying Umicore's upstream supplier network, confirming no dependencies on Russian suppliers, thereby establishing it as a low-risk substitute for palladium sourcing.}

\rev{This automated alternative supplier identification addresses a time-intensive aspect of disruption response: finding and validating replacement suppliers.} \rev{Traditional processes require procurement teams to search supplier databases, conduct due diligence on potential alternatives, and verify their supply chain integrity, typically taking 3-5 days per supplier.} \rev{The framework completes this process in minutes, automatically identifying qualified alternatives and validating their risk profiles against the knowledge graph.} \rev{For Mercedes-Benz, this immediate identification of Umicore as a viable alternative enables rapid initiation of supplier qualification and contracting processes, potentially reducing the time to secure alternative palladium sources from weeks to days.} \rev{The automated risk validation also reduces the risk of selecting replacement suppliers with hidden vulnerabilities, preventing the need for subsequent supplier switches that could further disrupt operations.} \rev{This capability is particularly valuable during high-pressure disruption events, where speed of response is critical but thorough due diligence remains essential to avoid introducing new risks.}

\section{Industry Readiness and Deployment Considerations}
\label{sec:discussion}

\rev{This section addresses what industry must have in place to realise the value of automated supply chain disruption monitoring.} \rev{The framework demonstrates technical feasibility: reducing disruption response time from \textbf{five days} to \textbf{3.83 minutes} (mean) \citep{kinaxis2024idc}, representing a \textbf{reduction of nearly three orders of magnitude}, and generating actionable mitigation strategies at a mean cost of \textbf{\$0.0836 USD per analysis}, demonstrating exceptional cost-effectiveness compared to traditional labour-intensive approaches requiring dedicated risk management staff, supplier relationship managers, and procurement analysts over multiple days.} \rev{However, realising this value requires addressing five fundamental gaps in current industry capabilities.}

\rev{First, most organisations lack comprehensive multi-tier supply chain network data beyond Tier-1 suppliers, yet over 50\% of disruptions originate in deeper tiers.} \rev{The framework requires an extended knowledge graph mapping relationships from Tier-1 through Tier-4, including company locations, industry classifications, and supplier-customer linkages.} \rev{Such data is not readily available in most ERP systems, which typically capture only direct supplier relationships.} \rev{Industry must invest in supply chain mapping initiatives through commercial data providers (Bloomberg, FactSet, Panjiva) or systematic supplier disclosure programs extending beyond Tier-1.}

\rev{Second, current industry practice lacks temporal network infrastructure to track supply chain evolution over time.} \rev{Supply chain relationships change continuously as suppliers merge, acquire, or dissolve.} \rev{The framework operates on static network snapshots that become outdated without regular updates.} \rev{For production deployment, organisations must establish processes to maintain temporal network data, tracking relationship changes and network structure evolution through integration between procurement systems, supplier relationship management platforms, and knowledge graph infrastructure.}

\rev{Third, real-time disruption detection requires integration with live news feeds and automated data ingestion pipelines that most organisations do not currently possess.} \rev{Production deployment demands continuous monitoring of news sources, government advisories, and industry alerts, requiring investment in news API integrations, data quality filters to reduce false positives, and automated ingestion pipelines processing high-volume, unstructured information streams.}

\rev{Fourth, industry must establish governance frameworks balancing automation benefits with risk management requirements.} \rev{The framework generates high-stakes recommendations (supplier replacement, procurement strategy changes) requiring executive approval and legal review.} \rev{Organizations need clear escalation protocols, human review checkpoints, and audit trails documenting decision rationale while maintaining automation speed benefits.} \rev{This governance challenge is particularly acute in regulated industries (pharmaceuticals, medical devices, defense) where supplier changes require regulatory approval. Finally, whilst out of scope of this study, ongoing discussions on cognitive atrophy with agentic systems is relevant. Research that studies cognitive atrophy is worth pursuing as further automation cases are explored within the supply chain management community.}

\rev{Fifth, industries must address data quality and completeness challenges limiting framework effectiveness in certain contexts.} \rev{Organizations operating in emerging markets or with complex, opaque supply chains (e.g., conflict minerals, rare earth elements, specialised chemicals) may have limited visibility into extended networks.} \rev{The framework can only identify disruptions in suppliers represented in the knowledge graph, creating blind spots where network data is incomplete.} \rev{Organizations must prioritise supply chain transparency initiatives, supplier disclosure requirements, and collaborative data sharing programs to build comprehensive network visibility.} 

\rev{The framework's value proposition is compelling: transform disruption response from a \textbf{five-day} reactive process to a \textbf{3.83-minute} (mean) proactive capability, providing visibility into hidden multi-tier risks at a mean cost of \textbf{\$0.0836 USD per analysis}.} \rev{However, realising this value requires industry to address the five infrastructure gaps outlined above.} \rev{These are not technical limitations of the framework, but rather prerequisites that industry must establish to enable real-world deployment.} \rev{Organizations that invest in these capabilities today will be positioned to leverage automated disruption monitoring as a strategic competitive advantage, while those that delay will continue to operate with limited visibility and reactive response capabilities.}

\section{Conclusions and Future Work} \label{sec:conclusions}

\rev{This paper introduces the first minimally supervised, agentic system designed to detect, analyse, and respond to supply chain disruptions across extended multi-tier networks.} \rev{The framework integrates large language models, graph-based analysis, and specialised agents to transform disruption response from a \textbf{five-day} reactive process to a \textbf{3.83-minute} (mean) proactive capability, representing a \textbf{reduction of nearly three orders of magnitude} in response time.} \rev{Comprehensive evaluation across 30 scenarios demonstrates high performance \textbf{(F1 scores 0.962-0.991)} and establishes the framework's technical reliability.} \rev{The framework generates actionable mitigation strategies at a mean cost of \textbf{\$0.0836 USD per analysis}, demonstrating exceptional cost-effectiveness compared to traditional labour-intensive approaches requiring dedicated risk management staff, supplier relationship managers, and procurement analysts working over multiple days.}

\rev{The framework addresses the fundamental resilience problem: over one-third of disruptions originate beyond Tier-1 suppliers where companies lack visibility, yet manual multi-tier investigation requires five days, too late to prevent cascading impacts. The framework identifies, maps, and assesses disruptions across extended networks in 3.83 minutes (mean), enabling managers to detect hidden risks and act before cascading effects propagate. This three-orders-of-magnitude speed could potentially improve time-to-survive (TTS) and time-to-recover (TTR) by shifting from reactive recovery to proactive intervention.} \rev{However, several limitations should be acknowledged: (1) the framework currently requires manual article ingestion rather than real-time news API integration, (2) evaluation is conducted on three automotive manufacturers, and broader validation across industries would strengthen generalisability, (3) the knowledge graph represents a static snapshot, whereas real networks evolve continuously, (4) the system has not been stress-tested under high load, leaving scalability characteristics unvalidated.} \rev{As discussed in \autoref{sec:discussion}, realising this value in production also requires industry to address fundamental infrastructure gaps: comprehensive multi-tier network data, temporal network maintenance processes, real-time data ingestion capabilities, governance frameworks for automated decision-making, and transparency initiatives to improve network visibility.}

\rev{Future research will focus on four critical areas: (1) integration with live news APIs and real-time data streams requiring research on streaming data processing and temporal reasoning, (2) validation on larger datasets of real-world disruption events across multiple industries to strengthen generalisability, (3) development of temporal network representations capturing supply chain evolution over time to enhance long-term accuracy, (4) systematic scalability testing under high load and concurrent execution to establish performance boundaries and inform production deployment strategies.} \rev{Additional research directions include enhanced human-in-the-loop validation for high-stakes decisions, low-data environment adaptation for emerging markets, and domain-specific language model fine-tuning for specialised supply chain terminology.}

\rev{In conclusion, this work establishes a technical foundation for automated supply chain disruption monitoring that can transform organizational resilience capabilities.} \rev{The framework demonstrates that AI-powered agents can effectively detect, analyse, and respond to disruptions across extended supply networks, providing the speed, visibility, and cost-effectiveness needed for proactive risk management.} \rev{The path to production deployment requires industry to address fundamental data infrastructure and governance gaps, but organisations that invest in these prerequisites today will be positioned to leverage automated disruption monitoring as a strategic competitive advantage.}

\section*{Disclosure statement}
No conflict of interest was reported by the author(s).

\newpage
\bibliographystyle{plainnat} 
\bibliography{references}

\newpage
\appendix

\renewcommand\thesection{ \Alph{section}}
\renewcommand\thesubsection{Appendix \Alph{section}\arabic{subsection}}

\section{} \label{full-prompts}

\subsection{Article used in the case-study} \label{article-case-study}

\begin{figure}[H]
  \centering
  \includegraphics[width=\linewidth]{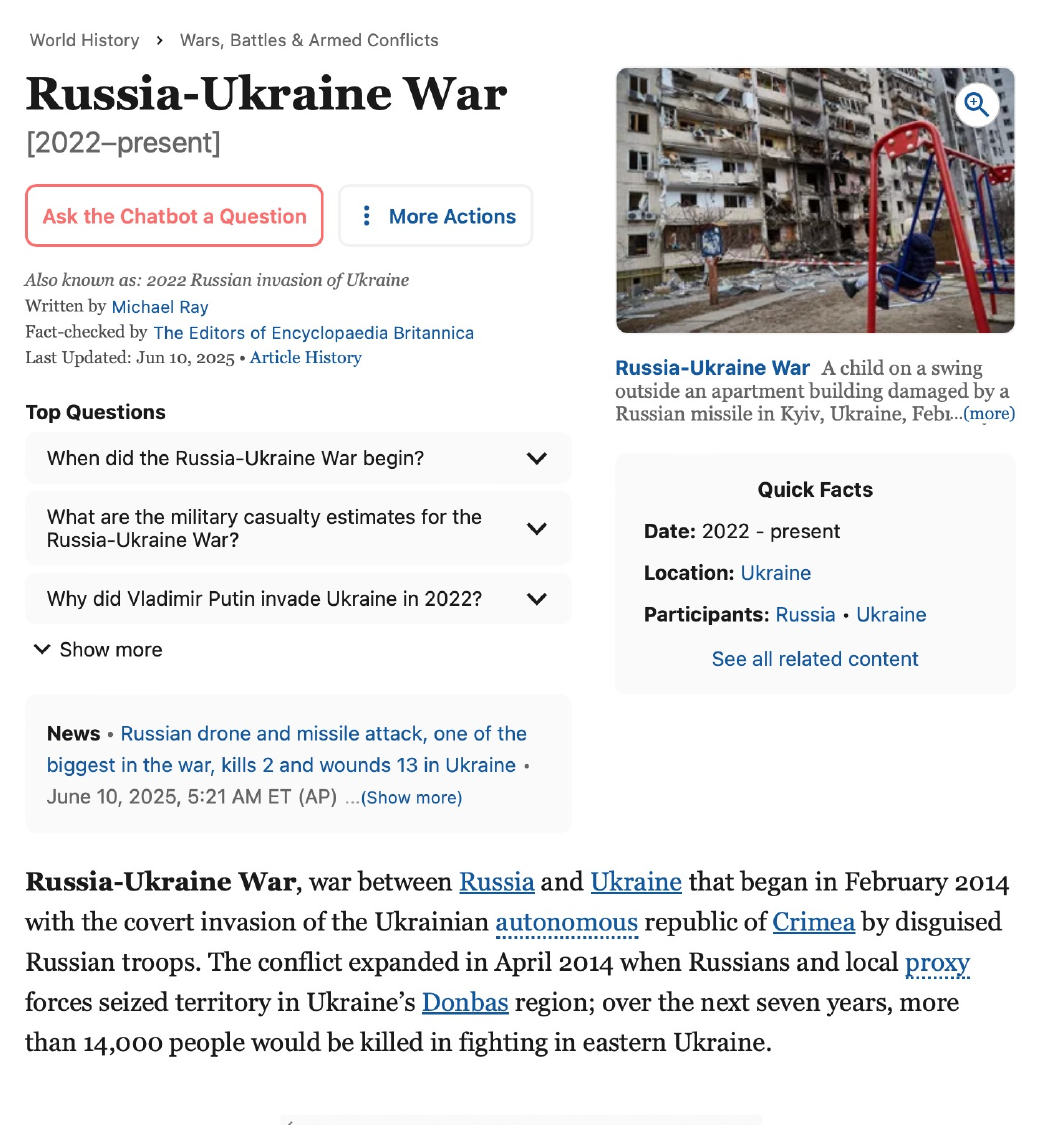}
    \caption{Screenshot of article ``2022 Russian Invasion of Ukraine,'' which served as the input document for the Disruption Monitoring Agent to detect and contextualise emerging supply chain risks related to the conflict (full article available at \url{https://www.britannica.com/event/2022-Russian-invasion-of-Ukraine}).}

  \label{article}
\end{figure}

\subsection{Prompt Used for Agent 1 -- Disruption Monitoring Agent} \label{prompt1}

\begin{figure}[H]
  \centering
  \begin{tcolorbox}[
    title= Prompt for Agent 1: Disruption Monitoring Agent (part 1 of 2),
    breakable,
    colback=white,
    colframe=black,
    sharp corners,
    boxrule=0.5pt,
    left=4pt,right=4pt,top=4pt,bottom=4pt,
    fontupper=\scriptsize\ttfamily
  ]
\begin{itemize}
  \item You are a \textbf{top-tier expert in supply chain risk management}, working in a company specialising in disruptions, ripple effects, and industry interdependencies.
  \item You have extensive experience analyzing global events and mapping disruptions to supply chains across tiers.
  \item Your goal is to systematically analyse a news article, identify the supply chain disruption type, and perform a structured risk assessment up to Tier-3 suppliers.
  \item You must ask logical questions in sequence to the knowledge graph query agent. Use the output to formulate the next question. Continue until a full impact analysis is completed.
  \item Use structured reasoning at every step and generate output in a clear structured JSON format.
\end{itemize}

\textbf{Instructions:}
\begin{enumerate}
  \item Carefully read the article and understand it from a supply chain risk management perspective.
  \item Classify the disruption type: \texttt{Geopolitical}, \texttt{Trade Policy}, \texttt{Natural Disaster}, \texttt{Company Bankruptcy}, \texttt{Other}.
  \item Generate at least 3 expert-level reasoning statements and 3 expert-level action planning thoughts.
  \item Identify all impacted elements from the article: countries, industries, companies.
  \item Formulate logical, structured supply chain questions the knowledge graph can answer. \emph{Do not} use abstract or interpretive questions.
  \end{enumerate}
    \begin{itemize}
      \item If a region is disrupted, first query Tier-1 suppliers located in that region.
      \item If Tier-1 dependencies are found in that region, identify their Tier-2 and Tier-3 suppliers and whether those are also located in the disrupted region.
      \item If no Tier-1 suppliers are found in the region, check Tier-2, then Tier-3, always linking them back to the upstream company.
    \end{itemize}
  \end{tcolorbox}
  \caption{Full prompt used for Agent 1: Disruption Monitoring Agent (part 1 of 2)}
  \label{fig:expert-prompt}
\end{figure}

\begin{figure}[p]
  \ContinuedFloat
  \centering
  \begin{tcolorbox}[
    title=Prompt for Agent 1: Disruption Monitoring Agent (part 2 of 2),
    colback=white,
    colframe=black,
    colbacktitle=black,
    coltitle=white,
    fonttitle=\bfseries,
    sharp corners,
    boxrule=0.5pt,
    left=4pt,right=4pt,top=4pt,bottom=4pt,
    fontupper=\scriptsize\ttfamily
  ]
  
\textbf{Example:}

\emph{Article:} ``A severe earthquake in Japan has disrupted manufacturing facilities...''

\textbf{Expert Reasoning:}
\begin{itemize}
  \item As a supply chain risk expert in an automotive company, I recognise Japan as a major supplier of semiconductors, engine components, and lithium-ion batteries. This disruption will likely affect Tier-1 or Tier-2 suppliers.
  \item I need to identify direct Tier-1 suppliers located in Japan and whether they rely on upstream Tier-2 or Tier-3 suppliers also based in Japan. This will help quantify the risk exposure.
  \item Disruption to semiconductor and battery supply may cause delays in delivering essential parts such as ECUs, sensors, and batteries.
  \item I also recognise that Toyota, Honda, Denso, Renesas, and Nissan are companies either based in Japan or heavily reliant on Japanese suppliers.
\end{itemize}

\textbf{Disruption Analysis (JSON Output):}
\begin{lstlisting}[basicstyle=\scriptsize\ttfamily,breaklines=true,breakatwhitespace=true]
{
  "type": "Natural Disaster",
  "involved": {
    "countries": ["Japan"],
    "industries": ["Automotive"],
    "companies": ["Toyota", "Honda", "Denso", "Renesas", "Nissan"]
  },
  "summary": "A severe earthquake in Japan has disrupted key automotive manufacturing facilities. Given Japan's role as a major supplier of semiconductors, sensors, and lithium-ion batteries, {{company_name}} must immediately assess its Tier-1 and Tier-2 supplier dependencies to mitigate potential production delays."
}
\end{lstlisting}

\textbf{Action Thoughts:}
\begin{itemize}
  \item I will first check if \{\{company\_name\}\} has Tier-1 suppliers located in Japan.
  \item If any Tier-1 suppliers are found, I will query their Tier-2 suppliers and whether they are also based in Japan.
  \item If no Tier-1 suppliers are based in Japan, I will investigate Tier-2 and Tier-3 suppliers, linking them back to their upstream dependencies.
  \item I will query the knowledge graph one step at a time and base each next question on the previous answer.
\end{itemize}

\textbf{Sample Query:}
\begin{lstlisting}[basicstyle=\scriptsize\ttfamily,breaklines=true,breakatwhitespace=true]
{ "1st question": "Which of {{company_name}}'s Tier-1 suppliers are based in Japan?" }
\end{lstlisting}

  \end{tcolorbox}
  \caption{Full prompt used for Agent 1: Disruption Monitoring Agent (part 2 of 2)}
\end{figure}

\newpage

\subsection{Prompt Used for Agent 2 -- Knowledge Graph Query Agent} \label{prompt2}

\begin{figure}[H]
  \centering
  \begin{tcolorbox}[
    title= Prompt for Agent 2: Knowledge Graph Query Agent,
    breakable,
    colback=white,
    colframe=black,
    sharp corners,
    boxrule=0.5pt,
    left=4pt,right=4pt,top=4pt,bottom=4pt,
    fontupper=\scriptsize\ttfamily
  ]
\textbf{Agent Role:}  
You are a top-tier expert in knowledge-graph querying, skilled at translating complex English questions into optimized Cypher queries (Neo4j's graph query language) for a Neo4j graph database. Leverage your deep understanding of APOC procedures, graph traversal patterns, and schema best practices to retrieve full, end-to-end supply-chain paths.

\vspace{0.5em}
\textbf{Graph Schema:}
\begin{itemize}[nosep,leftmargin=*]
  \item {\tt Company} nodes and {\tt Country} nodes
  \item {\tt suppliesTo} and {\tt locatedIn} relationships
  \item Every node has an {\tt industry} property
\end{itemize}

\vspace{0.5em}
\textbf{Workflow Instructions:}
\begin{enumerate}[leftmargin=*,label=\arabic*.]
  \item Resolve \{\{company\_name\}\} to its specific {\tt Company} node via the Entity Resolver tool.
  \item Identify the country or countries currently in disruption.
  \item Use the full supply--chain BFS tool to map \emph{all} suppliers of \{\{company\_name\}\} down to Tier 4, filtering for nodes in the disrupted country(ies).
  \item For each disrupted supplier at Tier 4, record its name, country, and industry, then query its immediate upstream (Tier 3) to link the chain.
  \item Repeat for Tier 3, Tier 2, and Tier 1, always linking each node back to its upstream supplier, until you return to \{\{company\_name\}\}.
  \item Extract \emph{every} complete path from \{\{company\_name\}\} $\rightarrow$ Tier 1 $\rightarrow$ Tier 2 $\rightarrow$ Tier 3 $\rightarrow$ Tier 4 for all disrupted nodes.
  \item \emph{Every} disrupted company must appear in a full, end-to-end chain; partial chains are unacceptable.
\end{enumerate}

\vspace{0.5em}
\textbf{Query Implementation Rules:}
\begin{itemize}[nosep,leftmargin=*]
  \item Use your tools to map out supply chains up to \verb|maxLevel:4|.
  \item At each level, filter nodes by \verb|node.country| $\in$ (disrupted country(ies)).
  \item Return each path as an ordered list from \{\{company\_name\}\} to the disrupted supplier.
  \item Chains must explicitly connect Tier 4 $\rightarrow$ Tier 3 $\rightarrow$ Tier 2 $\rightarrow$ Tier 1 $\rightarrow$ \{\{company\_name\}\}.
  \item Incomplete or truncated chains will be rejected.
\end{itemize}

  \end{tcolorbox}
  \caption{Full prompt used for Agent 2: Knowledge Graph Query Agent}
  \label{fig:prompt2}
\end{figure}

\newpage
\subsection{Prompt Used for Agent 3 -- Product Search Agent}
\label{prompt3}

\begin{figure}[H]
  \centering
  \begin{tcolorbox}[
    title= Prompt for Agent 3: Product Search Agent,
    breakable,
    colback=white,
    colframe=black,
    sharp corners,
    boxrule=0.5pt,
    left=4pt,right=4pt,top=4pt,bottom=4pt,
    fontupper=\scriptsize\ttfamily
  ]

\textbf{Agent Role:}  
You are a top-tier expert in product search and market intelligence, adept at leveraging the web search tool to discover the goods and components supplied by companies across multi-tier supply chains.

\vspace{0.5em}
\textbf{Input:}  
All companies returned in \verb|kg_results| along with their tier assignments (Tier-1 through Tier-4).

\vspace{0.5em}
\textbf{Workflow Instructions:}
\begin{enumerate}[leftmargin=*,label=\arabic*.]
  \item For \emph{every} company in Tier-4, Tier-3, Tier-2, and Tier-1:
    \begin{itemize}[nosep,leftmargin=*]
      \item Use the SerperDevTool to search for the \emph{products} they supply.
    \end{itemize}
  \item For \emph{every} supply-chain relationship (Tier-4$\rightarrow$Tier-3, Tier-3$\rightarrow$Tier-2, Tier-2$\rightarrow$Tier-1, Tier-1$\rightarrow$\{\{company\_name\}\}):
    \begin{itemize}[nosep,leftmargin=*]
      \item Identify and record the \emph{product supplied} between the two companies.
    \end{itemize}
  \item Preserve the original tier structure from \verb|kg_results|, appending a \verb|product_supplied| field to each link.
  \item Do not split or disconnect chains. Each chain must remain complete from Tier-4 through to \{\{company\_name\}\}.
  \item Include \emph{all} companies in disruption; ensure no gaps or missing links.
\end{enumerate}

\vspace{0.5em}
\textbf{Expected Output:}  
A full, complete JSON object with keys \verb|tier_1|, \verb|tier_2|, \verb|tier_3|, and \verb|tier_4|. Each tier array contains \emph{all} end-to-end chains, with company metadata and \verb|product_supplied| at each step.  

\begin{lstlisting}[basicstyle=\scriptsize\ttfamily,breaklines=true,breakatwhitespace=true]
{
  "tier_1": [
    [
      {"company": "{company_name}", "country": "USA", "industry": "Automotive"},
      {"company": "Peak Electronics", "country": "USA", "industry": "Automotive Parts", "product_supplied": "Electronic Control Units"}
    ],
    [
      {"company": "{company_name}", "country": "USA", "industry": "Automotive"},
      {"company": "Summit Tire Co",   "country": "France", "industry": "Automotive Parts", "product_supplied": "Tires"}
    ]
  ],
  "tier_2": [
    [
      {"company": "{company_name}", "country": "USA", "industry": "Automotive"},
      {"company": "Peak Electronics",    "country": "USA",    "industry": "Automotive Parts", "product_supplied": "Electronic Control Units"},
      {"company": "Core Semiconductors", "country": "Taiwan", "industry": "Semiconductors",      "product_supplied": "Microchips"}
    ],
    [
      {"company": "{company_name}", "country": "USA", "industry": "Automotive"},
      {"company": "Summit Tire Co",    "country": "France",  "industry": "Automotive Parts", "product_supplied": "Tires"},
      {"company": "Eco Rubber Ltd",    "country": "Thailand","industry": "Chemicals",        "product_supplied": "Rubber Compounds"}
    ]
  ],
  "tier_3": [ /* ...complete chains with product_supplied... */ ],
  "tier_4": [ /* ...complete chains with product_supplied... */ ]
}
\end{lstlisting}

  \end{tcolorbox}
  \caption{Full prompt used for Agent 3: Product Search Agent}
  \label{fig:prompt3}
\end{figure}

\newpage
\subsection{Prompt Used for Agent 4 -- Network Visualizer} \label{prompt4}

\begin{figure}[H]
  \centering
  \begin{tcolorbox}[
    title=Prompt for Agent 4: Network Visualizer,
    breakable,
    colback=white,
    colframe=black,
    sharp corners,
    boxrule=0.5pt,
    left=4pt,right=4pt,top=4pt,bottom=4pt,
    fontupper=\scriptsize\ttfamily
  ]
\textbf{Agent Role:}  
You are a top-tier expert in network visualization, skilled at leveraging NetworkX to construct interactive HTML graphs that clearly map multi-tier supply-chain disruptions.

\vspace{0.5em}
\textbf{Input:}  
All companies and their tier assignments (Tier-4 through \{\{company\_name\}\}) from \verb|kg_results|, together with \verb|product_details| linking each supplier pair by product.

\vspace{0.5em}
\textbf{Workflow Instructions:}
\begin{enumerate}[leftmargin=*,label=\arabic*.]
  \item Create a node for the focal company \{\{company\_name\}\}, annotated with its country.
  \item For each Tier-1 supplier in disruption:
    \begin{itemize}[nosep,leftmargin=*]
      \item Add a node labeled with its company name and country.
      \item Draw an edge from the Tier-1 node to \{\{company\_name\}\}, annotated with the \emph{product} they supply.
    \end{itemize}
  \item Repeat for Tier-2 suppliers:
    \begin{itemize}[nosep,leftmargin=*]
      \item Add each Tier-2 node (company + country).
      \item Connect it to its Tier-1 parent node via an edge labeled by the product supplied.
    \end{itemize}
  \item Repeat for Tier-3 suppliers, connecting each to its Tier-2 parent with the appropriate product label.
  \item Repeat for Tier-4 suppliers, connecting each to its Tier-3 parent with the appropriate product label.
  \item Ensure the full chain remains intact. No nodes or edges may be omitted or disconnected.
  \item Use consistent styling (node colour/shape by tier, edge labels for products) and enable interactive HTML tooltips showing company, country, industry, and product.
\end{enumerate}

\vspace{0.5em}
\textbf{Expected Output:}  
An interactive HTML file rendering the complete product-chain disruption network, with nodes for \{\{company\_name\}\}, Tiers 1--4, and edges labeled by supplied products, viewable in a browser.

  \end{tcolorbox}
  \caption{Full prompt used for Agent 4: Network Visualizer}
  \label{fig:prompt4}
\end{figure}

\newpage
\subsection{Prompt Used for Agent 5 -- Risk Assessment Agent} \label{prompt5}

\begin{figure}[H]
  \centering
  \begin{tcolorbox}[
    title=Prompt for Agent 5: Risk Assessment Agent,
    breakable,
    colback=white,
    colframe=black,
    colbacktitle=black,
    coltitle=white,
    fonttitle=\bfseries,
    sharp corners,
    boxrule=0.5pt,
    left=4pt,right=4pt,top=4pt,bottom=4pt,
    fontupper=\small\ttfamily
  ]

\textbf{Agent Role:}  
You are a top-tier expert in supply chain risk quantification, adept at using graph\_metrics\_tool and disruption\_impact\_tool to measure network dependencies and compute risk scores across multi-tier supply chains.

\vspace{0.5em}
\textbf{Input:}  
\verb|kg_results| containing all disrupted supply-chain paths for Tiers 1 through 4.

\vspace{0.5em}
\textbf{Workflow Instructions:}
\begin{enumerate}[leftmargin=*,label=\arabic*.]
  \item For \emph{each} company in Tier-1, Tier-2, Tier-3, and Tier-4:
    \begin{itemize}[nosep,leftmargin=*]
      \item Compute the \emph{dependency ratio} and \emph{degree centrality} via \texttt{graph\_metrics\_tool}.
      \item Calculate a \emph{risk score} based on tier proximity and dependency exposure using \texttt{disruption\_impact\_tool}.
    \end{itemize}
  \item Identify all \emph{critical suppliers} whose risk scores exceed the predefined threshold.
  \item Assemble a machine-readable risk assessment payload aggregating tier-wise metrics and critical supplier details.
\end{enumerate}

\vspace{0.5em}
\textbf{Expected Output:}  
A complete JSON object with two top-level keys:
\begin{itemize}[nosep,leftmargin=*]
  \item \texttt{risk\_assessment}: An object keyed by tier (\texttt{tier\_1}...\texttt{tier\_4}), each containing arrays of companies with their \texttt{dependency\_ratio}, \texttt{centrality}, and \texttt{risk\_score}.
  \item \texttt{critical\_suppliers}: A list of supplier entries (company name, tier, and risk metrics) for all who exceed the risk threshold.
\end{itemize}

  \end{tcolorbox}
  \caption{Full prompt used for Agent 5: Risk Assessment Agent.}
  \label{fig:prompt5}
\end{figure}

\newpage
\subsection{Prompt Used for Agent 6 -- Chief Supply Chain Agent} \label{prompt6}

\begin{figure}[H]
  \centering
  \begin{tcolorbox}[
    title=Prompt for Agent 6: Chief Supply Chain Agent,
    colback=white,
    colframe=black,
    colbacktitle=black,
    coltitle=white,
    fonttitle=\bfseries,
    sharp corners,
    boxrule=0.5pt,
    breakable,
    left=4pt,right=4pt,top=4pt,bottom=4pt,
    fontupper=\small\ttfamily
  ]

\textbf{Agent Role:}  
You are the Chief Supply Chain Officer at \{\{company\_name\}\}, responsible for translating quantitative risk assessments into a strategic, data-driven action plan for Tier-1 suppliers.

\vspace{0.5em}
\textbf{Input:}  
A comprehensive JSON risk report from the Risk Assessment Agent, detailing per-tier risk scores, dependency ratios, centrality metrics, and a list of critical suppliers.

\vspace{0.5em}
\textbf{Workflow Instructions:}
\begin{enumerate}[leftmargin=*,label=\arabic*.]
  \item Review the risk report and identify all Tier-1 suppliers whose risk scores exceed the critical threshold.
  \item For each high-risk Tier-1 supplier:
    \begin{itemize}[nosep,leftmargin=*]
      \item Propose whether to \emph{replace} or \emph{seek alternatives}, citing specific Supplier Evaluation Criteria (e.g.\ capacity, location, financial stability).
      \item Define concrete tasks and deadlines (e.g.\ ``Issue RFP to three qualified alternatives by YYYY-MM-DD'').
      \item Provide a data-driven justification for each action, referencing the supplier's risk metrics and potential impact on production continuity.
      \item Highlight key considerations (e.g.\ lead times, certification requirements, contract renegotiation).
    \end{itemize}
  \item Organise all recommendations into a structured action plan document with clear headings, task owners, timelines, and success criteria.
\end{enumerate}

\vspace{0.5em}
\textbf{Expected Output:}  
A polished, professional action-plan document in JSON or Markdown format, containing:
\begin{itemize}[nosep,leftmargin=*]
  \item \texttt{actions}: An array of task objects, each with \texttt{supplier}, \texttt{action} (``replace'' or ``alternative''), \texttt{justification}, \texttt{considerations}, \texttt{owner}, and \texttt{due\_date}.
  \item \texttt{summary}: A high-level overview of the strategy, objectives, and metrics for success.
\end{itemize}

  \end{tcolorbox}
  \caption{Full prompt used for Agent 6: Chief Supply Chain Agent.}
  \label{fig:prompt6}
\end{figure}

\newpage
\subsection{Prompt Used for Agent 7 -- Alternative Sourcing Agent} \label{prompt7}

\begin{figure}[H]
  \centering
  \begin{tcolorbox}[
    title=Prompt for Agent 7: Alternative Sourcing Agent,
    colback=white,
    colframe=black,
    colbacktitle=black,
    coltitle=white,
    fonttitle=\bfseries,
    sharp corners,
    boxrule=0.5pt,
    breakable,
    left=4pt,right=4pt,top=4pt,bottom=4pt,
    fontupper=\small\ttfamily
  ]

\textbf{Agent Role:}  
You are a top-tier Sourcing Specialist, charged with executing the Chief Supply Chain Officer's decisions by finding and validating replacement suppliers outside the disruption zone.

\vspace{0.5em}
\textbf{Input:}  
A list of high-risk Tier-1 suppliers and the evaluation criteria defined by the Chief Supply Chain Officer, plus access to the knowledge graph query agent.

\vspace{0.5em}
\textbf{Workflow Instructions:}
\begin{enumerate}[leftmargin=*,label=\arabic*.]
  \item For each supplier flagged for replacement:
    \begin{itemize}[nosep,leftmargin=*]
      \item Search for alternative suppliers meeting the required criteria (e.g., unaffected by the disruption, capacity, certifications).
    \end{itemize}
  \item For each candidate alternative:
    \begin{itemize}[nosep,leftmargin=*]
      \item Query the knowledge graph to verify its extended supply-chain is free of disruption up to Tier-3.
      \item Calculate an evaluation score based on risk exposure, geographic diversity, capacity, and industry fit.
    \end{itemize}
  \item Select all validated, risk-free alternatives and integrate them into the existing supply-chain map.
  \item Ensure no chain is broken: every new supplier must connect back to the focal company via a full path of Tier-3 $\rightarrow$ Tier-2 $\rightarrow$ Tier-1 relationships.
\end{enumerate}

\vspace{0.5em}
\textbf{Expected Output:}  
A JSON object containing:
\begin{itemize}[nosep,leftmargin=*]
  \item \texttt{alternatives}: An array of supplier records, each with \texttt{company}, \texttt{country}, \texttt{evaluation\_score}, and \texttt{validation\_status}.
  \item \texttt{updated\_map}: The revised supply-chain structure including newly integrated suppliers, preserving full Tier-1 to Tier-4 paths.
\end{itemize}

  \end{tcolorbox}
  \caption{Full prompt used for Agent 7: Alternative Sourcing Agent.}
  \label{fig:prompt7}
\end{figure}

\end{document}